\documentclass{article}

\PassOptionsToPackage{numbers,sort,compress}{natbib}

\usepackage[preprint]{neurips_2025}

\usepackage[toc,page,header]{appendix}
\usepackage{minitoc}

\doparttoc

\usepackage{graphicx}
\usepackage{lipsum}
\usepackage{listings}
\usepackage[table]{xcolor}
\usepackage{array}
\usepackage{tabularx}    
\usepackage{ragged2e}    
\newcolumntype{P}[1]{>{\RaggedRight\arraybackslash}p{#1\linewidth}}
\usepackage{algorithm2e}

\usepackage[utf8]{inputenc} 
\usepackage[T1]{fontenc}    
\usepackage{hyperref}       
\usepackage{url}            
\usepackage{booktabs}       
\usepackage{amsfonts}       
\usepackage{nicefrac}       
\usepackage{microtype}      
\usepackage{amsmath}
\usepackage{bm}
\usepackage{enumitem}
\usepackage{subcaption}
\usepackage{wrapfig}

\newcommand{\nop}[1]{}

\title{CALM: Co-evolution of Algorithms and Language Model for Automatic Heuristic Design}

\author{
Ziyao Huang\textsuperscript{1}, Weiwei Wu\textsuperscript{2}, Kui Wu\textsuperscript{3}, Jianping Wang\textsuperscript{1}, Wei-Bin Lee\textsuperscript{4}\\[0.1cm]
\textsuperscript{1}City University of Hong Kong\;\textsuperscript{2}Southeast University\\
\textsuperscript{3}University of Victoria\;\textsuperscript{4}Hon Hai Research Institute\\[0.1cm]
\texttt{zhuang88-c@my.cityu.edu.hk, weiweiwu@seu.edu.cn, wkui@uvic.ca}\\
\texttt{jianwang@cityu.edu.hk, wei-bin.lee@foxconn.com}
}

\begin{document}

\maketitle


\begin{abstract}
  Tackling complex optimization problems often relies on expert-designed heuristics, typically crafted through extensive trial and error. Recent advances demonstrate that large language models (LLMs), when integrated into well-designed evolutionary search frameworks, can autonomously discover high-performing heuristics at a fraction of the traditional cost. However, existing approaches predominantly rely on verbal guidance, i.e., manipulating the prompt generation process, to steer the evolution of heuristics, without adapting the underlying LLM. We propose a hybrid framework that combines verbal and numerical guidance, the latter achieved by fine-tuning the LLM via reinforcement learning based on the quality of generated heuristics. This joint optimization allows the LLM to co-evolve with the search process. Our method outperforms state-of-the-art (SOTA) baselines across various optimization tasks, running locally on a single 24GB GPU using a 7B model with INT4 quantization. It surpasses methods that rely solely on verbal guidance, even when those use significantly more powerful API-based models.
\end{abstract}

\section{Introduction}\label{sec:introduction}

Complex optimization problems are prevalent in real-world applications, including logistics~\cite{duan2022data, tresca2022automating}, scheduling~\cite{mihoubi2021reactive, palacio2022q}, and transportation~\cite{dahmani2024solving, pereira2021r5r}. Traditionally, solving these problems relies heavily on manually crafting high-quality heuristics, a labor-intensive process requiring substantial expert knowledge. Given the limitations of this manual approach, Automatic Heuristic Design (AHD) emerged to streamline heuristic generation. Nevertheless, classic AHD approaches like Genetic Programming (GP)~\cite{burke2009exploring} still depend significantly on human-defined problem-specific components, limiting the search space and flexibility. 

Recently, the advent of Large Language Models (LLMs) has introduced promising avenues for AHD by employing LLMs as heuristic generators and evolutionary computing (EC) techniques as a search framework. In this paradigm, heuristics generated by LLMs are iteratively evaluated through a predefined simulation framework, and superior heuristics inform subsequent generation prompts, thus creating a feedback-driven evolutionary loop~\cite{eoh}. Nevertheless, existing LLM-based AHD methods predominantly keep the underlying LLM untouched and merely guide heuristic evolution via textual prompt manipulations, referred to as "verbal gradients"~\cite{reevo}. Consequently, these methods inherently neglect the opportunity of tuning and enhancing the generative capability of LLM based on the feedback from heuristic designs. 

We propose Co-evolution of Algorithms and the Language Model (CALM) to capture this opportunity. CALM fundamentally differs from the state-of-the-art~\cite{eoh,reevo,hsevo,mcts-ahd} by enabling the LLM to co-evolve alongside heuristic designs. This co-evolution is made possible by treating the heuristic generation process not only as a target of optimization but also as a rich source of training data. As heuristics are continually proposed, evaluated, and selected based on their performance, the evolutionary loop naturally produces abundant prompt-response-performance triplets. These data points are highly informative, as each heuristic's effectiveness provides an implicit signal about the utility of the underlying generation process. By using this signal as feedback for reinforcement learning, we can fine-tune the LLM, thereby applying what we term "numerical gradients" to adapt the model itself. This co-evolution approach unlocks a new dimension of adaptability, allowing the LLM to internalize characteristics of successful heuristics and improve its future generations\nop{ for specific tasks}.

CALM is the first LLM-based AHD framework that jointly optimizes both the prompt generation process and the LLM model itself, overcoming the limitations of fixed-model approaches. For prompt generation, CALM introduces a suite of evolutionary operators, including fine-granularity mutation operators (injection and replacement) and a diversity-aware crossover operator, that promote meaningful and diverse heuristic variations while preserving structural coherence. Furthermore, a simple yet effective collapse mechanism is developed to help escape the local optima. For model improvement, CALM employs a memory-efficient reinforcement learning (RL) algorithm GRPO~\cite{grpo} with a carefully designed reward function to enable efficient fine-tuning. Experimental results demonstrate that our new approach can discover heuristics that beat existing state-of-the-art baselines~\cite{eoh,reevo,mcts-ahd}, while running entirely on a local computer with a single 24GB GPU, in contrast to prior methods that depend heavily on commercial LLM APIs.

\begin{figure*}
    \centering
    \includegraphics[width=0.998\linewidth]{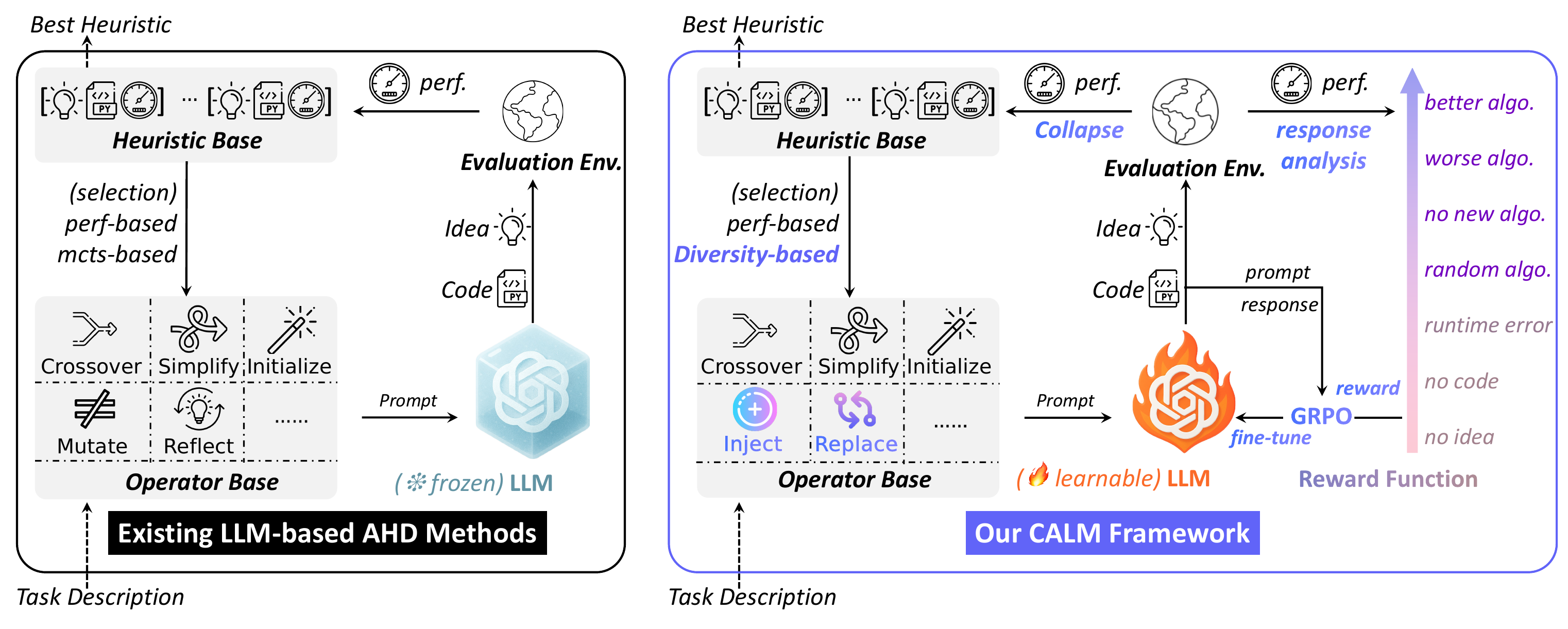}
    \caption{Pipeline of existing LLM-based AHD methods~\cite{funsearch,reevo,hsevo,mcts-ahd} under a fixed LLM and our new approach CALM that enables the co-evolution of LLM in the iterative heuristic search process. New components are presented in bright colors.}
    \label{fig:intro}
    \vspace{-0.5cm}
\end{figure*}

\section{Related Work}\label{sec:related-work}
As our approach centers on fine-tuning LLMs by RL for solving optimization problems, we review relevant literature in both RL and LLMs applied to optimization. Additional related topics, including LLMs for code generation and RL-based LLM fine-tuning, are discussed in Appendix~\ref{appendix-sec:related-work}.

\textbf{RL for Optimization Problems.}
Existing RL-based methods for optimization can be broadly categorized by the role the learned policy plays:
\textit{\textbf{(1) Instance-Level Solution Generator.}}
Deep RL has been widely adopted to learn policies for solving specific optimization instances~\cite{kwon2020pomo,pan2023adjustable,bi2024learning}. However, these methods differ fundamentally from LLM-based AHD methods, as they directly produce solutions rather than design the algorithms that generate them. The LLM-based AHD approach operates at \textit{a meta level}, seeking to learn the algorithmic structure that produces solutions. This distinction also applies to the broader class of Neural Combinatorial Optimization (NCO)~\cite{luo2024self,xiao2024neural,sui2024neuralgls,zheng2023pareto}, where models are trained to directly solve instances. Moreover, NCO methods often require explicit adaptation to handle problem scales not seen during training, whereas our method generalizes more naturally to new scales.
\textbf{\textit{(2) Heuristic Generator.}}
Some RL-based methods target meta-level search to discover heuristics instead of instance-level solutions. For example, AlphaDev~\cite{mankowitz2023faster} learns to combine low-level operations to discover faster sorting algorithms, and~\cite{yi2022automated} searches for high-performing metaheuristics from predefined algorithmic components. While having similar goals, these approaches rely heavily on hand-engineered building blocks, akin to traditional AHD frameworks~\cite{pillay2018hyper,sanchez2021feature,burke2009exploring}. In contrast, LLM-based method reduces manual intervention by leveraging LLMS to explore an open-ended heuristic space with minimal prior specification.

\textbf{LLM for Optimization Problems.}
Studies in this area fall into two categories depending on how LLMs are employed:
\textbf{\textit{(1) Instance-Level Solution Generator.}}
Several works~\cite{abgaryan2024llms,jiang2024unco,wu2024netllm} prompt LLMs with instance-specific inputs for direct solution generation. LLM-based methods in this category focus on discovering reusable heuristics.\nop{, which are more fundamental and transferable.} Moreover, methods such as~\cite{jiang2024unco,wu2024netllm} keep LLM parameters frozen, and~\cite{abgaryan2024llms} fine-tunes the model using supervised labels from an existing solver (OR-Tools~\cite{ortools}). In contrast, our method fine-tunes the LLM solely by providing it with "how good is your current response" instead of "how should you do", encouraging a more flexible exploration in heuristic generation.
\textit{\textbf{(2) Heuristic Generator.}}
LLM-based AHD methods~\cite{liu2023algorithm,qube,funsearch,eoh,reevo,liu2024llm4ad,hsevo,mcts-ahd,alphaevolve} repeatedly ingest information about the current elite heuristics—typically their natural-language descriptions, source code, and performance scores—and, via fixed prompt templates that mimic genetic operators, produce new candidate heuristics. Those candidates are then executed and evaluated, and the resulting feedback is fed back into the prompt, forming an evaluate–generate loop that continues until the evaluation budget is exhausted. However, prior work keeps the LLM static. Our approach improves this by continuously fine-tuning the LLM using prompt-response-performance tuples from the evolutionary process, enhancing future heuristic generation.
Notably, a very recent study EvoTune~\cite{evotune} also fine-tunes LLMs within an evolutionary search framework, but our method differs in four key ways: (i) We extra adapt prompt generation to better guide the LLM's evolution; (ii) We continuously reward the model by telling it how good each response is to encourage open-ended exploration, while EvoTune offers a binary feedback by indicating the better of two responses; (iii) Our method, using a compact LLM, outperforms several mainstream approaches~\cite{funsearch, eoh, reevo, hsevo, mcts-ahd} that rely on more powerful LLMs. In contrast, EvoTune only reports improvements over~\cite{funsearch} when both use compact LLMs.

\section{Preliminary}

\subsection{LLM-Based AHD}
Let $P$ be a problem with input space $\mathcal{I}$ and solution space $\mathcal{S}$, and let a \emph{heuristic} be a function $h:\mathcal{I}\to\mathcal{S}$.  
Given a training set $D\subset\mathcal{I}$ and an objective $f:\mathcal{S}\to\mathbb{R}$ (lower is better), the performance of a heuristic is
$g(h)=\mathbb{E}_{x\in D}\!\left[-f\!\left(h(x)\right)\right]$.
Let $\mathcal{H}$ denote the space of all feasible heuristics. The objective of AHD is to identify the optimal heuristic within this space, i.e., 
$h^{*}\;=\;\arg\max_{h\in \mathcal{H}} g(h).$

LLM-based AHD is AHD where LLM serves as a heuristic generator. In practice, the LLM is charged with designing the core decision function of a solver. For example, on tasks like the Traveling Salesman Problem (TSP) or the Capacitated Vehicle Routing Problem (CVRP), an LLM-based AHD method might generate a function, which selects the next city to visit or constructs an edge-desirability matrix to guide solution search within an Ant Colony Optimization (ACO) framework.

\subsection{GRPO}
GRPO~\cite{grpo} is a recent reinforcement learning algorithm that has proven effective in training LLMs, as evidenced by its application in models such as DeepSeek-R1. GRPO starts from an initial model $\pi_\theta$ and a reward function denoted by $r_\phi(q,o)$ that maps the prompt $q$ and the generated response $o$ to a scalar. At the beginning of each training round, it snapshots $\pi_\theta$ as a reference model $\pi_\mathrm{ref}$. Then, it split all task prompts into multiple batches. When training for each prompt batch $\mathcal{D}_b$, it first snapshots $\pi_\theta$ as $\pi_\mathrm{old}$. For each task prompt $q\in\mathcal{D}_b$, it samples a group of $G$ responses $\{o_i\}_{i=1}^G\sim\pi_{\theta_\mathrm{old}}$ and computes rewards $\bm{r}=\{r_i=r_\phi(q, o_i)\}_{i=1}^G$ for each prompt-response pair. Subsequently, it computes the advantage $\hat{A}_{i,t}$ for each token $t$ in response $i$ as the normalized reward $(r_i-\mathrm{mean}(\bm{r})) / \mathrm{std}(\bm{r})$. The model parameters $\theta$ are updated by maximizing the following objective function:
{
\begin{multline}\label{eq:grpo_obj}
\mathcal J_{\mathrm{GRPO}}(\theta)
=\mathbb E_{\left[q\sim\mathcal Q,\{o_i\}\sim\pi_{\theta_{\mathrm{old}}}\right]}\\
\frac1G\sum_{i=1}^G\frac1{|o_i|}\sum_{t=1}^{|o_i|}\left\{\min\left[\hat r_{i,t}\hat A_{i,t},\;\mathrm{clip}\left(\hat r_{i,t},1-\varepsilon,1+\varepsilon\right)\hat A_{i,t}\right]-\beta\,\mathbb{D}_{\mathrm{KL}}\bigl[\pi_\theta\|\pi_{\mathrm{ref}}\bigr]\right\},
\end{multline}
}
\noindent where $\epsilon$ and $\beta$ are hyper-parameters, $\hat{r}_{i,t}={\pi_\theta(o_{i,t}\mid q,o_{i,<t})}/{\pi^{\mathrm{old}}_\theta(o_{i,t}\mid q,o_{i,<t})}$,
\nop{\begin{equation}
    \hat{r}_{i,t}=\frac{\pi_\theta(o_{i,t}\mid q,o_{i,<t})}{\pi^{\mathrm{old}}_\theta(o_{i,t}\mid q,o_{i,<t})},
\end{equation}}
and the KL divergence term is computed using the unbiased estimator in~\cite{schulman2020approximating} with guaranteed positivity.
\nop{
\begin{equation}
    \mathbb{D}_{KL} \left[ \pi_{\theta} \| \pi_{\text{ref}} \right] = 
\frac{\pi_{\text{ref}}(o_{i,t} \mid q, o_{i,<t})}{\pi_{\theta}(o_{i,t} \mid q, o_{i,<t})}
- \log \frac{\pi_{\text{ref}}(o_{i,t} \mid q, o_{i,<t})}{\pi_{\theta}(o_{i,t} \mid q, o_{i,<t})}
- 1
\end{equation}
}
GRPO uses the group mean reward as a baseline to eliminate the need for an auxiliary value network, thereby reducing memory requirements. Additionally, the clipping mechanism combined with KL divergence regularization ensures stable and conservative updates.

\section{Methodology}

To explore the benefit of RL-based fine-tuning for discovering higher-quality heuristics in LLM-based AHD, we introduce CALM, a novel framework that integrates both verbal and numerical guidance in evolutionary heuristic search.
As shown in Fig.~\ref{fig:intro}, CALM maintains a pool of heuristics, each with its own idea, code, and performance. At every round, CALM draws a feasible evolutionary operator to produce a new prompt $q$. Subsequently, $G$ responses are sampled from the local LLM $\pi_\theta$, which are then evaluated. Based on the evaluation results, rewards are assigned to each response for GRPO to train the LLM, and new feasible heuristics are added to the pool. Consequently, CALM returns the best-so-far heuristic after running $T$ rounds.
Next, we elaborate on the critical techniques in CALM: prompt generation, collapse mechanism, and the reward function.


\subsection{Prompt Generation}
CALM provides several evolutionary operators: injection, replacement, crossover, simplification, and initialization. Prompts are predominantly generated by the selected operator and heuristics sampled from maintained pools. The initialization operator is an exception, as it does not require heuristics from the pool. A complete algorithm can be found in Appendix~\ref{appendix-sec:algorithm}. Next, we introduce the heuristic sampling method and operators in detail\footnote{The prompts used in CALM are detailed in Appendix~\ref{appendix-sec:prompts} due to space limit.}.

\textbf{Heuristic Sampling Method.} The heuristic sampling approach varies for the crossover operator, details of which will be provided when introducing this operator. For the remaining operators, i.e., injection, replacement, and simplification, the heuristics are selected based on their performance rankings like~\cite{eoh}. Specifically, the probability of sampling a heuristic \( h \) is inversely proportional to its rank in the current pool (i.e., proportional to \( 1/\mathrm{rank}_{p}(h) \)). Heuristics ranked below a threshold, defined as the population size, are assigned a probability of zero.

\textbf{Fine-Granularity Mutation Operators: Injection \& Replacement.} GRPO assigns an advantage score to each token based on the relative reward of the full response compared to others from the same prompt. This means each part of a heuristic is encouraged or penalized depending on the quality of the whole. However, heuristic performance can shift dramatically with changes to even a single sub-component, making uniform treatment of all parts—in terms of gradient direction—unreliable.

While cumulative gradient updates can correct misattributed rewards or penalties for the same token appearing in different responses, we aim to further boost this process. To this end, we introduce two novel operators that enable more precise control over heuristic variations. These operators encourage the LLM to retain more common parts while introducing meaningful modifications to the input heuristic. Consequently, GRPO is expected to more effectively identify the contribution—positive or negative—of individual structural changes. The two newly designed operators are:

    \textbf{\textit{Injection.}} Given an existing heuristic, the injection operator prompts the LLM to incorporate a new component into it. Additionally, a concise description of the new component must be included in the response. All component descriptions are stored, and subsequent applications of the injection operator require the LLM to introduce components distinct from those previously saved, promoting diversity in generated heuristics. Unlike mutation operators in prior LLM-based AHD methods~\cite{mcts-ahd, eoh}, which are fed with full heuristic codes: (1) Our approach uses compact summaries instead of full code, allowing more heuristics to fit within the LLM’s context window; (2) Saved component descriptions are globally accessible and not limited to the currently sampled heuristics; (3) Prior methods often require entirely new heuristics, while our approach focuses on more granular modifications; (4) When the number of heuristics is below the population size, the sampling probability of the injection operator is increased to encourage exploration in the phase of population expansion.
    
    \textbf{\textit{Replacement.}} Given an existing heuristic, the replacement operator prompts the LLM to rewrite an existing component under a specific instruction. There are three distinct instructions, and each time the replacement operator is applied, one is randomly sampled for the given heuristic. While the "rewrite hyper-parameter" instruction is also present in prior studies~\cite{eoh, mcts-ahd}, CALM introduces two novel instructions: (1) Rewrite an instance-independent decision rule as an instance-dependent one—to improve the heuristic’s adaptability to varying problem contexts; (2) Rewrite a fragment that assigns equal or near-equal credit to all candidates as one that differentiates credit based on contextual performance—to encourage more effective prioritization and refined decision-making.

\paragraph{Diversity‐Aware Crossover.}
To balance exploitation and exploration, each crossover invocation randomly chooses between
(1) \emph{performance‐based}: sample both parents by performance rank; and
(2) \emph{diversity‐based}: sample the first parent \(h_{c,1}\) by performance rank and the second from all retained heuristics with probability inversely proportional to diversity rank (larger diversity is better). Specifically, let \(\mathrm{idea\_token}(\cdot)\) denote the set of unique tokens in a heuristic's idea, the diversity is: $\mathrm{div}(h_{c,1},h)
  = \lvert\mathrm{idea\_token}(h)\setminus \mathrm{idea\_token}(h_{c,1})\rvert/\lvert\mathrm{idea\_token}(h)\rvert.$
\nop{
\begin{equation}
      \mathrm{div}(h_{c,1},h)
  = \frac{\lvert\mathrm{idea\_token}(h)\setminus \mathrm{idea\_token}(h_{c,1})\rvert}{\lvert\mathrm{idea\_token}(h)\rvert},
\end{equation}
}
This hybrid mechanism ensures that at least one parent heuristic is of high quality, while the second parent is either high-performing or structurally novel. The diversity-aware selection expands the evolutionary search space and leverages underutilized heuristics, potentially unlocking novel strategies that might otherwise be overlooked due to suboptimal early performance. More discussions are moved to Appendix~\ref{appendix-sec:crossover}.

\textbf{Simplification Operator.} As heuristic structures grow increasingly complex through repeated applications of injection, crossover, and replacement, there is a risk of accumulating redundant or unnecessarily verbose components. The simplification operator counterbalances this tendency by prompting the LLM to produce a more concise and effective version of a given heuristic.

\textbf{Initialization Operator.} In cases where there is no heuristic in the pool (e.g., no initial/seeding function is provided), this operator is invoked to prompt the LLM to generate new heuristics.

\subsection{Collapse Mechanism}

\textbf{Why to Collapse.} A key reason LLM-based evolutionary heuristic search can succeed is that prompts containing better-performing heuristics tend to guide the LLM toward generating even stronger ones. This creates a self-reinforcing feedback loop, gradually evolving a population of increasingly effective heuristics. However, this process can also lead to inbreeding and premature convergence: over time, the population becomes dominated by minor variations of the current best-performing heuristic. When this state persists without meaningful breakthroughs, the search risks becoming trapped in a local optimum, a classic challenge in evolutionary computing~\cite{Eshelman1991PreventingPC}.

\textbf{How to Collapse.} As a remedy, CALM introduces a proactive collapse mechanism that resets the search process when it detects stagnation, allowing the system to escape local optima and reinitiate meaningful exploration. Specifically, when the search has plateaued—characterized by a prolonged lack of performance improvement—we reset the population by discarding all heuristics except two: the original seed algorithm and the current best-performing heuristic. These two retained heuristics jointly serve as the seed algorithms for the new search process, grounding it in past progress while freeing it from the genetic redundancy accumulated in the previous population.

\textbf{When to Collapse.} 
Once the heuristic pool reaches its target population size, CALM begins tracking stagnation using a no-breakthrough counter \( c_n \), initialized to zero. This counter records the number of consecutive prompt rounds—each involving \( G \) sampled responses—that fail to yield a globally superior heuristic. If any sampled heuristic in a round surpasses all previous ones in performance, \( c_n \) is reset to zero; otherwise, it increments by one.

To escape local optima, CALM introduces a probabilistic collapse mechanism based on this counter. At the end of each round, collapse is triggered if:
$\mathrm{random}(0, 1) < c_n \delta_0 \quad \text{or} \quad c_n \geq C,$
where \( \delta_0 \ll 1 \) controls the rate at which collapse probability grows, and $C$ is a hard cap ensuring collapse happens by the $C$-th stagnation step at the latest. 
To aid in hyperparameter selection, we further provide an analytical approximation for the expected number of rounds before collapse is triggered:
\begin{equation}\label{eq:collapse-round-expectation}
  \mathbb{E}\left[c_n \mid \mathrm{collapse},C>\frac{1}{\delta_0}\right]
  \;\approx\;\sqrt{\frac{\pi}{2\delta_0}}.
\end{equation}

This collision of a rising‐probability rule with a fixed maximum fosters a balance between giving the search plenty of room to improve and ensuring it doesn’t stall infinitely. A detailed proof and discussion about the benefit of the mechanism can be found in Appendix~\ref{appendix-sec:collapse}.

\subsection{Reward Function}\label{sec:reward}

When the LLM is prompted to generate a heuristic, the response quality varies considerably. CALM’s reward function is designed to provide fine-grained guidance for GRPO-based learning by penalizing infeasible outputs and reinforcing improvements over existing heuristics. 

\textbf{Infeasible Responses.} The most penalized outcomes are those where the generated response fails to constitute a valid heuristic. To diagnose such cases, we apply a hierarchy of failure modes, assigning progressively higher (i.e., less negative) rewards to increasingly plausible but still unacceptable outputs. These modes include: (1) omission of a required idea (reward: $-1.0$); (2) missing code block ($-0.95$); (3) improperly formatted function ($-0.9$); (4) runtime errors or time budget violations ($-0.85$); and (5) detection of randomness in the heuristic ($-0.75$), which incurs the mildest penalty among infeasible cases. The reward for detecting randomized components, denoted as $r_\mathrm{rand} < 0$, also serves as a ceiling for all other infeasibility penalties.%
This graded reward design encourages the LLM to progress toward syntactic and semantic correctness even if full feasibility is not yet achieved.

\textbf{Feasible Responses.} For valid heuristics, performance becomes the primary signal for learning. However, since the quality of generated heuristics is strongly influenced by the prompt---particularly the included base heuristics---we avoid attributing full credit or blame for the resulting performance solely to the LLM. Therefore, rewards depend not only on the absolute performance of the new heuristic but also on its novelty and improvement relative to the best base heuristic in the prompt. Specifically, let $H$ denote the set of base heuristics used to construct prompt $q$, and $h_\mathrm{new}$ be the heuristic parsed from the LLM’s output $o$. We define the top base heuristic as $h_\mathrm{t\_base} = \arg\max_{h \in H} g(h)$, and measure relative performance via:
\begin{equation}
    \Delta(h_\mathrm{new}, h_\mathrm{t\_base}) = \mathrm{clip}\left( \frac{|g(h_\mathrm{new}) - g(h_\mathrm{t\_base})|}{\min\{|g(h_\mathrm{new})|, |g(h_\mathrm{t\_base})|\}}, 0, 1 \right).
\end{equation}

The reward function $r_\phi(q, o \mid h_\mathrm{new}, h_\mathrm{t\_base})$ is then defined as:
\begin{equation}\label{eq:reward}
    r_\phi(q, o \mid h_\mathrm{new}, h_\mathrm{t\_base}) =
    \begin{cases}
        0.8 r_\mathrm{rand}, & \text{if } \exists h \in H \text{ s.t. } g(h) = g(h_\mathrm{new}); \\
        0.5 r_\mathrm{rand} \cdot \Delta(h_\mathrm{new}, h_\mathrm{t\_base}), & \text{if } g(h_\mathrm{new}) < g(h_\mathrm{t\_base}); \\
        1 + \Delta(h_\mathrm{new}, h_\mathrm{t\_base}), & \text{if } g(h_\mathrm{new}) > g(h_\mathrm{t\_base}).
    \end{cases}
\end{equation}

This formulation satisfies two critical properties. First, any feasible heuristic is rewarded more than any infeasible one, ensuring that basic correctness is prioritized. Second, the reward is primarily determined by whether the new heuristic improves over the best base heuristic or not, with the relative performance gap further modulating the strength of the reward or penalty. When the generated heuristic is identical in performance to an existing base heuristic, a small but consistent reward ($0.8 r_\mathrm{rand}$) is given to discourage trivial reproduction. If the new heuristic underperforms relative to the best base, a scaled negative reward is applied, while genuine improvements yield strictly positive rewards starting from 1. In the special case of heuristics generated via the initialization operator (which has no base references), the reward is simply set to zero to encourage feasible outputs.

\section{Experiments}\label{sec:experiment}

\textbf{Implementation Details of CALM.} 
We build CALM on Unsloth~\cite{unsloth} and employ an INT4-quantized Qwen2.5-7B-Instruct model \cite{yang2024qwen2}, fine-tuning just 1.15\% of its weights. INT4 compression cuts memory usage up to 8× versus FP32 but degrades precision. According to \cite{yang2024qwen2}, performance ranks as follows: GPT-4o-mini $\approx$ Qwen2.5-Turbo > Qwen2.5-14B-Instruct > Qwen2.5-7B-Instruct > Qwen2.5-7B-Instruct-INT4. The 14B and 7B Instruct models share the same architecture, so the larger parameter count drives the 14B’s edge over the 7B, while quantization further reduces the 7B’s accuracy. Consequently, GPT-4o-mini–based baselines retain a clear advantage in raw accuracy over our lean, resource-efficient setup. More implementation details can be found in Appendix~\ref{appendix-sec:running-time}

\textbf{Optimization Tasks.}
Existing LLM-based methods can demonstrate near-optimal or optimal performance on some benchmark problems, such as TSP~\cite{eoh, reevo, mcts-ahd} (aided by ACO solvers) and knapsack problem (KP)~\cite{mcts-ahd}, leaving little room for further improvement. Therefore, we focus on tasks that remain challenging for LLM-based AHD as follows: Online Bin Packing (OBP) problem and TSP under step-by-step construction task, CVRP and Orienteering Problem (OP) under an ACO search framework. Detailed problem descriptions can be found in Appendix~\ref{appendix-sec:problem}
	

\textbf{Baselines.} To evaluate CALM, we compare its designed heuristics against the following baselines: (1) hand-crafted heuristics such as Best-Fit~\cite{kenyon1995best} for OBP, Greedy-Construct (GC)~\cite{rosenkrantz1977analysis} for TSP, and ACO~\cite{blum2005ant} for CVRP and OP; (2) Nerual Combinatorial Optimization (NCO) methods including POMO~\cite{kwon2020pomo} and DeepACO~\cite{ye2023deepaco}; and (3) LLM-based AHD approaches like FunSearch~\cite{funsearch}, EoH~\cite{eoh}, ReEvo~\cite{reevo}, HSEvo~\cite{hsevo}, and MCTS-AHD~\cite{mcts-ahd}. To ensure a fair comparison, we align CALM and all LLM-based AHD baselines with consistent settings, including shared seed heuristics, identical training datasets for evaluating heuristic performance, and comparable evaluation budgets--specifically, 1,000 heuristic evaluations\nop{\footnote{In prior LLM-based AHD approaches~\cite{reevo, mcts-ahd}, a cap is usually placed on the number of successful heuristic evaluations. One such evaluation typically consumes multiple LLM queries. For instance, in~\cite{mcts-ahd}, only responses that present a well-structured heuristic idea and produce new (i.e., previously unseen) code are counted as valid evaluations. Hence, our budget of $500 \times 4 = 2000$ queries corresponds to approximately $1000$ evaluations.}} for baselines and a fixed budget of 2,000 LLM queries for CALM across all tasks except OBP. Notably, prior AHD methods typically conduct 2,000 heuristic evaluations using over 4,000 queries for OBP, whereas CALM operates under a fixed budget of 2,000 queries.

\subsection{Overall Results}
\vspace{-0.3cm}
\begin{table}[ht]
\centering
\caption{Optimality gaps of construction heuristics for OBP, averaged over three runs. All methods use the same training and test datasets as~\cite{mcts-ahd}. Gaps are computed against a strong lower bound~\cite{martello1990lower}. Test instances follow the Weibull distribution, with scales either matching (underlined) or differing from training. Format: 1k\_100 denotes instances with $1,000$ items and a bin capacity of $100$.}
\label{tab:obp}
\resizebox{.7\linewidth}{!}{
\begin{tabular}{lccccccc}
\toprule
\multicolumn{8}{c}{\textbf{Online Bin Packing (OBP)}} \\
\midrule
Test sets  
  & \underline{1k\_100}
  & \underline{1k\_500}
  & \underline{5k\_100}
  & \underline{5k\_500}
  & 10k\_100
  & 10k\_500
  & Avg. \\
\midrule
Best Fit~\cite{kenyon1995best}   & 4.77\% & 0.25\% & 4.31\% & 0.55\% & 4.05\% & 0.47\% & 2.40\% \\
First Fit  & 5.02\% & 0.25\% & 4.65\% & 0.55\% & 4.36\% & 0.50\% & 2.56\% \\
\midrule
\multicolumn{8}{c}{\emph{LLM‐based AHD:} \textbf{GPT-4o-mini} (w/o. GRPO)} \\
\midrule
FunSearch~\cite{funsearch}  & \cellcolor{gray!25}\textbf{2.45\%} & 0.66\% & 1.30\% & 0.25\% & 1.05\% & 0.21\% & 0.99\% \\
EoH~\cite{eoh}        & 2.69\% & 0.25\% & 1.63\% & 0.53\% & 1.47\% & 0.45\% & 1.17\% \\
ReEvo~\cite{reevo}      & 3.94\% & 0.50\% & 2.72\% & 0.40\% & 2.39\% & 0.31\% & 1.71\% \\
HSEvo~\cite{hsevo}      & 2.64\% & 1.07\% & 1.43\% & 0.32\% & 1.13\% & 0.21\% & 1.13\% \\
MCTS-AHD~\cite{mcts-ahd}   & \cellcolor{gray!25}\textbf{2.45\%} & 0.50\% & 1.06\% & 0.32\% & 0.74\% & 0.26\% & 0.89\% \\
CALM (Ours) & 2.78\% & 0.29\% &\cellcolor{gray!25}\textbf{0.83\%} & 0.28\% &\cellcolor{gray!25}\textbf{0.50\%}  &0.24\% & 0.82\%\\
\midrule
\multicolumn{8}{c}{\emph{LLM‐based AHD:} \textbf{Phi3.5-mini-Instruct(w/. DPO, Over $10\times$ LLM Queries)}} \\
\midrule
EvoTune~\cite{evotune}    & 4.77\%  & 0.25\% & 4.27\% & 0.40\% & 4.05\%  & 0.42\% & 2.36\% \\
\midrule
\multicolumn{8}{c}{\emph{LLM‐based AHD:} \textbf{Qwen2.5-7B-Instruct-INT4 (w/. GRPO)}} \\
\midrule
CALM (Ours)& 2.55\% 
           & \cellcolor{gray!25}\textbf{0.00\%} 
           & 0.85\% 
           & \cellcolor{gray!25}\textbf{0.17\%} 
           & 0.56\%
           & \cellcolor{gray!25}\textbf{0.14\%} 
           & \cellcolor{gray!25}\textbf{0.71\%} \\
\bottomrule
\end{tabular}}
\end{table}

\begin{wraptable}[23]{r}{0.5\textwidth}
  \centering
  \vspace{-0.44cm}
  \caption{Performance of step-by-step construction heuristics on TSP, averaged over three runs. All methods are evaluated on three test sets of 1,000 instances each, using the same training and testing datasets as in~\cite{mcts-ahd}. Test sets with in-domain scales (i.i.d.\ to the training data) are underlined. Optimal tour lengths are obtained by LKH~\cite{lin1973effective}. The best LLM-based result in each column is shaded gray, and the overall best result is shown in bold.}
  \resizebox{\linewidth}{!}{%
    \begin{tabular}{l cc cc cc}
      \toprule
      \multicolumn{7}{c}{\textbf{Traveling Salesman Problem (TSP)}} \\
      \midrule
       & \multicolumn{2}{c}{\underline{N=50}}
       & \multicolumn{2}{c}{N=100} 
       & \multicolumn{2}{c}{N=200} \\
      \cmidrule(lr){2-3}\cmidrule(lr){4-5}\cmidrule(lr){6-7}
      Methods            & Obj.$\downarrow$ & Gap$\downarrow$     & Obj.$\downarrow$ & Gap$\downarrow$     & Obj.$\downarrow$ & Gap$\downarrow$     \\
      \midrule
      Optimal            & 5.675     & --        & 7.768     & --        & 10.659    & --        \\
      GC~\cite{rosenkrantz1977analysis}   & 6.959     & 22.62\%   & 9.706     & 24.94\%   & 13.461    & 26.29\%   \\
      POMO~\cite{kwon2020pomo}               & \textbf{5.697} & \textbf{0.39\%} & \textbf{8.001} & \textbf{3.01\%} & 12.897    & 20.45\%   \\
      \midrule
      \multicolumn{7}{c}{\emph{LLM-based AHD:} \textbf{GPT-3.5-turbo (w/o. GRPO)}} \\
      \midrule
      FunSearch~\cite{funsearch}          & 6.683     & 17.75\%   & 9.240     & 18.95\%   & 12.808    & 19.61\%   \\
      EoH~\cite{eoh}                & 6.390     & 12.59\%   & 8.930     & 14.96\%   & 12.538    & 17.63\%   \\
      MCTS-AHD~\cite{mcts-ahd}          & 6.346     & 11.82\%   & 8.861     & 14.08\%   & 12.418    & 16.51\%   \\
      \midrule
      \multicolumn{7}{c}{\emph{LLM-based AHD:} \textbf{GPT-4o-mini (w/o. GRPO)}} \\
      \midrule
      FunSearch~\cite{funsearch}          & 6.357     & 12.00\%   & 8.850     & 13.93\%   & 12.372    & 15.54\%   \\
      EoH~\cite{eoh}                & 6.394     & 12.67\%   & 8.894     & 14.49\%   & 12.437    & 16.68\%   \\
      MCTS-AHD~\cite{mcts-ahd}           & \cellcolor{gray!25}6.225     & \cellcolor{gray!25}9.69\%  & 8.684     & 11.79\%   & 12.120    & 13.71\%   \\
      CALM (Ours)        & 6.273     & 10.54\%   & 8.691  & 11.88\%     &   12.104    & 13.56\%   \\
      \midrule
      \multicolumn{7}{c}{\emph{LLM-based AHD:} \textbf{Qwen2.5-7B-Instruct-INT4 (w.\ GRPO)}} \\
      \midrule
      CALM (Ours)        & 6.244     & 10.04\%   
                         & \cellcolor{gray!25}8.668   & \cellcolor{gray!25}11.58\%  
                         & \cellcolor{gray!25}\textbf{12.088}  & \cellcolor{gray!25}\textbf{13.41\%} \\
      \bottomrule
    \end{tabular}%
  }
  \label{tab:tsp}
\end{wraptable}

\textbf{OBP.}
We train and evaluate CALM on the same dataset used in~\cite{mcts-ahd}, which includes four training instances at varying scales and five testing instances spanning six different scales—two of which are out-of-domain and not seen during training. Results in Table~\ref{tab:obp} show that CALM consistently outperforms all baseline methods in terms of average optimality gap across the full test set. It can achieve superior performance on out-of-domain and in-domain scales. Remarkably, CALM achieves a zero gap in set 1k\_500, indicating exact optimal solutions at that scale.

As EvoTune~\cite{evotune} does not release its code, training procedure, or test data, we adapted its reported heuristic to our context for comparison. We used the hyperparameter optimization framework Optuna~\cite{optuna} to tune all hyperparameters over 2000 trials, selecting the best configuration based on the average optimality gap across all test sets. While EvoTune outperforms some prior RL-free LLM-based baselines on certain test sets, our method consistently achieves better results than EvoTune across all test sets. Furthermore, according to EvoTune’s reported data, their approach improves FunSearch by 2.88\%\textasciitilde15.29\% under the same LLM on OBP. In contrast, our CALM framework outperforms FunSearch by 38.69\% (averaged over the performance improvements across all test sets). These results highlight the greater efficiency and robustness of our LLM-based AHD framework.

\begin{table}[ht]
  \centering
  \caption{Performance of ACO-based heuristics on CVRP and OP, averaged over three runs. All methods are evaluated on three test sets of 64 randomly generated instances each, following the setup in~\cite{mcts-ahd} and~\cite{hsevo}, respectively. Optimal solutions are approximated using DeepACO with significantly more ants and iterations than those in the baseline configurations.}
  \label{tab:cvrp_op}
  \resizebox{0.95\linewidth}{!}{%
    \begin{tabular}{l cc cc cc cc cc cc}
      \toprule
      & \multicolumn{6}{c}{\textbf{CVRP}} & \multicolumn{6}{c}{\textbf{OP}} \\
      \midrule
      & \multicolumn{2}{c}{\underline{N=50}} & \multicolumn{2}{c}{N=100} & \multicolumn{2}{c}{N=200}
      & \multicolumn{2}{c}{\underline{N=50}} & \multicolumn{2}{c}{N=100} & \multicolumn{2}{c}{N=200} \\
      \cmidrule(lr){2-3}\cmidrule(lr){4-5}\cmidrule(lr){6-7} \cmidrule(lr){8-9}\cmidrule(lr){10-11}\cmidrule(lr){12-13}
      Methods      & Obj.$\downarrow$ & Gap$\downarrow$ & Obj.$\downarrow$ & Gap$\downarrow$ & Obj.$\downarrow$ & Gap$\downarrow$
                   & Obj.$\uparrow$ & Gap$\downarrow$ & Obj.$\uparrow$ & Gap$\downarrow$ & Obj.$\uparrow$ & Gap$\downarrow$ \\
      \midrule
      Optimal      & 8.888 & –       & 14.932 & –       & 27.159 & –
                   & 19.867 & –       & 36.392 & –       & 63.380 & –       \\
      ACO          & 18.581 & 109.05\% & 30.107 & 101.63\% & 37.590 & 40.69\%
                   & 13.354 & 32.69\%  & 24.131 & 33.69\%  & 37.586 & 40.69\%  \\
      \midrule
      \multicolumn{13}{c}{\emph{LLM-based AHD:} \textbf{GPT-4o-mini (w/o. GRPO)}} \\
      \midrule
      EoH~\cite{eoh}          & 9.894  & 11.32\%  & 16.953 & 13.54\%  & 30.314 & 11.62\%
                   & 13.388 & 32.61\%  & 24.154 & 33.63\%   & 37.319 & 41.12\% \\
      ReEvo~\cite{reevo}        & 9.558  & 7.54\%   & 16.350 & 9.50\%   & 29.219 & 7.58\%
                   & \cellcolor{gray!25}\textbf{15.103} & \cellcolor{gray!25}\textbf{23.98\%} & 30.523 & 16.13\% & 53.807 & 15.10\% \\
      HSEvo~\cite{hsevo}        & 9.431  & 6.11\%   & 16.396 & 9.81\%   & 29.520 & 8.69\%
                   & 15.082 & 24.08\%  & 30.454 & 16.32\%   & 53.260 & 15.97\% \\
      MCTS-AHD~\cite{mcts-ahd}     & 9.372  & 5.44\%   & 15.974 & 6.98\%   & 28.434 & 4.70\%
                   & 14.847 & 25.27\%  & 30.163 & 17.12\%   & 53.024 & 16.34\% \\
      CALM (Ours)  & 9.404  & 5.81\%   & 16.046 & 7.46\%   & 28.713 & 5.72\%
                   & 15.017 & 24.41\%  & 30.294 & 16.76\%   & 53.098 & 16.22\% \\
      \midrule
      \multicolumn{13}{c}{\emph{LLM-based AHD:} \textbf{Qwen2.5-7B-Instruct-INT4 (w/. GRPO)}} \\
      \midrule
      CALM (Ours)  & \cellcolor{gray!25}\textbf{9.228} & \cellcolor{gray!25}\textbf{3.83\%} & \cellcolor{gray!25}\textbf{15.745} & \cellcolor{gray!25}\textbf{5.44\%} & \cellcolor{gray!25}\textbf{28.230} & \cellcolor{gray!25}\textbf{3.95\%}
                   & 15.054 & 24.22\%  & \cellcolor{gray!25}\textbf{30.78} & \cellcolor{gray!25}\textbf{15.43\%} & \cellcolor{gray!25}\textbf{55.406} & \cellcolor{gray!25}\textbf{12.58\%} \\
      \bottomrule
    \end{tabular}%
  }
\vspace{-0.1cm}
\end{table}

\textbf{TSP.} CALM is trained on the same dataset used by \cite{mcts-ahd}: a training set of 64 TSP instances with $N=50$ nodes and three test sets of 1,000 instances each at $N=50,100$, and $200$. As shown in Table~\ref{tab:tsp}, CALM-constructed heuristics outperform all LLM-based baselines on both out-of-domain test sets and achieve the second-best LLM-based result on the in-domain set. Notably, at the largest scale, CALM surpasses the NCO baseline POMO, which requires per-scale training. 

\textbf{CVRP.} CALM is trained on 10 instances as in~\cite{mcts-ahd} with $N=50$ nodes using the ACO framework, and evaluated on three test sets of 64 instances each at $N=50, 100$, and $200$, following the same generation protocol. During both training and testing, the number of ants and iterations is fixed to 30 and 100, respectively. As shown in Table~\ref{tab:cvrp_op}, CALM consistently outperforms all LLM-based baselines across all test sets, including the in-domain one and both out-of-domain ones.

\textbf{OP.} CALM is trained 5 OP instances with $N=50$ nodes using the ACO framework and evaluated on three test sets of 64 instances each at $N=50, 100$, and $200$, following the generation protocol in~\cite{mcts-ahd}. Both training and testing use a fixed configuration of 20 ants and 50 iterations. As reported in Table~\ref{tab:cvrp_op}, CALM consistently outperforms all other LLM-based baselines on the out-of-domain scales. As for the in-domain scale, it still outperforms EoH and the most recent approach MCTS-AHD.

\subsection{Discussion}

\textbf{Efficacy of our verbal gradient.}
For each problem instance, we further evaluate the design of our verbal gradient in isolation (i.e., without GRPO) by (1) switching the backend to the GPT-4o-mini API, (2) setting \(G=1\), and (3) using \(T=4000\) for OBP and \(T=2000\) for all other tasks—matching the query budgets of prior LLM-based AHD methods. As shown in Tables~\ref{tab:obp}--\ref{tab:cvrp_op}, this API-based variant of CALM delivers performance on par with or superior to the recent MCTS-AHD approach: it achieves the lowest optimality gaps on the 5k\_100 and 10k\_100 OBP datasets and ranks second on average across all OBP test sets, matches MCTS-AHD and outperforms all other baselines on every CVRP test set, consistently surpasses MCTS-AHD on all OP instances, and closely tracks MCTS-AHD on TSP at \(N=50\) and \(100\) while outperforming all non-MCTS baselines at those scales and even surpassing MCTS-AHD at \(N=200\). These results demonstrate that, \emph{even without reinforcement learning or advanced techniques such as reflection~\cite{reevo,hsevo} and tree search~\cite{mcts-ahd}, CALM’s verbal guidance mechanism remains highly effective, placing the API-based CALM firmly within the top tier of existing LLM-based AHD methods.}

\begin{wrapfigure}[34]{r}{0.5\textwidth}
  \centering

  \begin{minipage}{\linewidth}
    \vspace{-0.45cm}
    \centering
    \begin{subfigure}[b]{0.48\linewidth}
      \includegraphics[width=\linewidth]{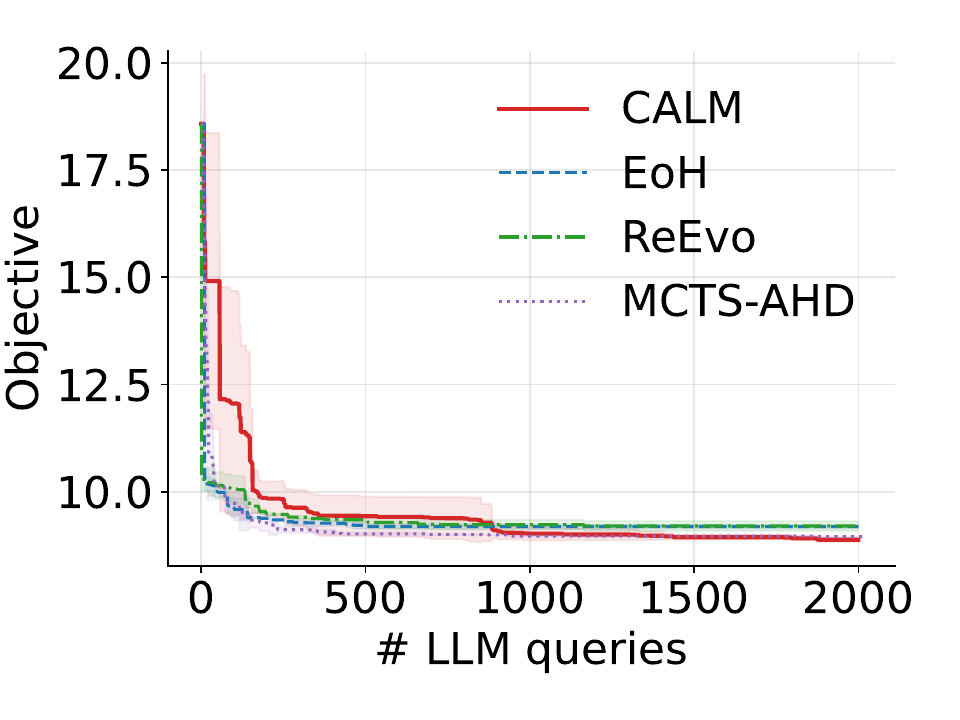}
      \caption{CVRP.}\label{fig:a}
    \end{subfigure}\hfill
    \begin{subfigure}[b]{0.48\linewidth}
      \includegraphics[width=\linewidth]{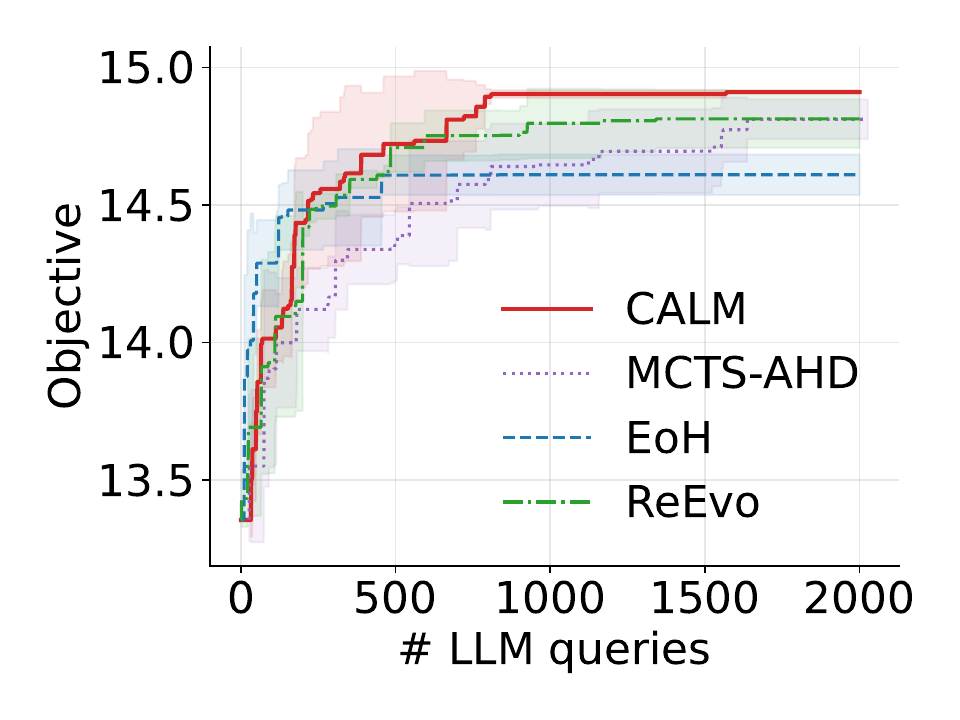}
      \caption{OP.}\label{fig:b}
    \end{subfigure}
    \caption{Objective score of the best heuristic in training averaged over 3 runs (std.\ dev.\ shaded).}
    \label{fig:training-curve}
  \end{minipage}

  \vspace{0.75\baselineskip}

  \begin{minipage}{\linewidth}
    \centering
    {\footnotesize        
    \captionof{table}{Optimality gap under ablation settings for problem OBP and OP averaged over three runs.}
    \label{tab:ablation}
    \setlength{\tabcolsep}{4pt} 
    \begin{tabular}{lrr}
      \toprule
      \textbf{Method}                & \textbf{OBP} & \textbf{OP} \\
      \midrule
      CALM (local, w/ GRPO)          & 0.71\%  & 17.41\% \\
      CALM (API, w/o GRPO)           & 0.82\%  & 19.13\% \\
      \midrule
      \multicolumn{3}{l}{\textit{RL-based Fine-tuning}} \\
      local, w/o GRPO                & 1.78\%  & 19.89\% \\
      rew$\in\{0.5r_\mathrm{rand},1\}$& 1.04\%  & 17.44\% \\
      rew=performance                & 1.24\%  & 21.30\% \\
      \midrule
      \multicolumn{3}{l}{\textit{Collapse Mechanism}}\\
      w/o Collapse                    & 0.98\%  & 19.57\% \\
      $\delta_0=0.0005$, $C=15$       & 0.77\%  & 18.31\% \\
      $\delta_0=0.005$,  $C=15$       & 1.93\%  & 27.22\% \\
      $\delta_0=0.0005$, $C=\infty$   & 0.96\%  & 19.50\% \\
      $\delta_0=0.005$,  $C=\infty$   & 0.98\%  & 18.38\% \\
      \midrule
      \multicolumn{3}{l}{\textit{Operators}}\\
      w/o diversity                   &  1.05\%  & 19.44\% \\
      w/o crossover                   & 0.88\%  & 18.49\% \\
      w/o injection                   & 1.11\%  & 18.68\% \\
      w/o replacement                 & 1.20\%  & 17.57\% \\
      w/o simplification              & 1.35\%  & 19.45\% \\
      \bottomrule
    \end{tabular}
    } 
  \end{minipage}
\end{wrapfigure}

\textbf{Power of RL.}
We have tested the performance of CALM without the GRPO algorithm and under many ablation settings. As shown in Table~\ref{tab:ablation}, results demonstrate that disabling the GRPO module causes the largest drop in performance across near all ablations. In other words, \emph{The reinforcement‐learning component has the most significant impact on overall performance among all ablation settings}.
Morever, as illustrated in Table~\ref{tab:obp}\textasciitilde\ref{tab:cvrp_op}, \emph{with GRPO and our custom reward, the Qwen2.5-7B-Instruct-INT4–derived heuristic not only closes the gap but actually outperforms the GPT-4o-mini–based heuristic.}
We have also visualized the training curve in Figure~\ref{fig:training-curve}. Results show CALM’s heuristics lag early—likely due to GPT-4o-mini’s head start—but as GRPO adapts the LLM, its heuristics converge and outperform all baselines. This suggests the transformative power of RL in enhancing AHD.

\textbf{Impact of reward design.}
Our feasible‐response reward allocates credit by comparing each generated heuristic against its parent(s), rather than attributing full reward or blame solely to the LLM. We evaluate two alternative schemes (keeping the infeasible‐response penalty unchanged): (i) \emph{performance‐based reward}, where a feasible heuristic receives a positive reward proportional to its performance relative to the seed algorithm; and (ii) the \(\{0.5\,r_{\mathrm{rand}},\,1\}\)\nobreakdash‐\emph{improvement reward}, which assigns reward \(1\) if the new heuristic outperforms all parent or baseline heuristics, and \(0.5\,r_{\mathrm{rand}}\) otherwise. Both alternatives remove the trivial‐reproduction penalty and mitigate the performance bias present in Equation~(\ref{eq:reward}). As Table~\ref{tab:ablation} demonstrates, neither variant beats our original design: the performance‐based scheme underperforms even the no‐RL baseline on the OP problem, while the \(\{0.5\,r_{\mathrm{rand}},\,1\}\)-improvement strategy delivers closer but still inferior results compared to our proposed reward function. This confirms the effectiveness of our original reward design.

\textbf{Impact of collapse.}
We examine the impact of the collapse mechanism by analyzing the heuristics produced by CALM both without collapse and under various hyperparameter configurations that influence when collapse is triggered. As shown in Table~\ref{tab:ablation}, incorporating the collapse mechanism generally enhances the heuristic search process. An exception arises in the configuration with the strictest tolerance for not discovering a breakthrough heuristic (i.e., when $\delta_0=0.005$ and $C=15$). A detailed analysis of the evolutionary trajectory under this setting reveals a significantly reduced number of breakthroughs. In one run on the OP problem, no breakthrough heuristic was identified after the 132nd LLM query. These findings suggest that setting a reasonable tolerance for the absence of breakthroughs—balancing patience with the benefits of early stopping—is important for supporting a more effective evolution.

\textbf{Impact of operators.}
We evaluate each operator’s contribution by measuring CALM’s performance with that operator removed (Table~\ref{tab:ablation}). Results show all operators positively impact heuristic quality. Crossover, injection, and replacement are similarly critical—removing any one notably degrades performance in either OBP or OP. Among all, removing simplification causes the largest drop in both tasks, likely because it uniquely reduces redundancy and curbs complexity, counterbalancing other operators that tend to increase heuristic length. Moreover, when crossover is applied without diversity-based selection—using only performance-based sampling—CALM performs worse than with no crossover at all, highlighting the importance of diversity awareness in the most-used operator.

\section{Conclusion}\label{sec:conclusion}
This paper introduces CALM, the first framework to marry prompt evolution with on-the-fly LLM adaptation for AHD, freeing it from the constraints of fixed-model approaches. Running entirely on a single 24 GB GPU with a compact foundation model, CALM autonomously uncovers heuristics that outmatch SOTA API-based baselines across various challenging optimization scenarios. Moreover, even without the power of RL, CALM matches or exceeds prior best results using the same LLM API, demonstrating the potency of our verbal-gradient designs. In the future, we expect that scaling CALM’s paradigm to larger models and extended post-training could further push the frontier of automated algorithm discovery.

\bibliography{ref.bib}
\bibliographystyle{unsrtnat}

\newpage
\appendix

\section{Extended Discussion about Related Work}\label{appendix-sec:related-work}

\textbf{LLM for Code Generation.}
Recent work has explored improving LLMs’ code generation capabilities through post-training~\cite{islam2024llm,tsai2024code,wang2024enhancing,shen2024policy,li2024fine}. For example,\cite{islam2024llm} employs RL and semantic feedback to repair vulnerabilities, while\cite{wang2024enhancing} demonstrates RL’s effectiveness in enhancing code quality. Despite surface similarities, our task differs fundamentally: in code generation, objectives often prioritize pass rates~\cite{shen2024policy,wang2024enhancing,tsai2024code} or safety~\cite{li2024fine,islam2024llm}, whereas our goal is to maximize heuristic performance. Moreover, in code generation, fine-tuning aims to produce a generally stronger model, while in our case, both the model tuning and prompt evolution serve a singular goal—improving the quality of generated heuristics.

\textbf{RL for LLM Fine-tuning.}
Reinforcement learning is a central technique for fine-tuning large language models, with the RLHF paradigm commonly relying on Proximal Policy Optimization (PPO)\cite{schulman2017proximal} to iteratively refine model outputs based on human feedback. Building on this, Group Relative Policy Optimization (GRPO)\cite{grpo} simplifies training by removing the need for a separate value network, instead estimating baselines over groups of candidate completions—leading to improved sample efficiency and stability. Other alternatives such as Direct Preference Optimization (DPO), SLiC-HF~\cite{zhao2023slic}, and Rejection Sampling Optimization (RSO)\cite{liu2023statistical} offer off-policy mechanisms that further reduce computational burden. While we do not aim to develop new fine-tuning algorithms, our method integrates GRPO within the broader co-evolution framework to adapt the LLM in tandem with heuristic evolution. We specifically adopt GRPO because it requires only a scalar signal per prompt-response pair (in contrast to preference-based signals), making it suitable for our setting. Moreover, we implement fine-tuning using Unsloth\cite{unsloth}, a GPU-efficient open-source framework that enables fast, memory-light training even on single consumer-grade GPUs—making our method especially practical and accessible for researchers with limited hardware resources.

\section{Complete Algorithm}\label{appendix-sec:algorithm}
\RestyleAlgo{ruled}
\begin{algorithm}[H]
    \SetKwInOut{Input}{Input}
    \SetKwInOut{Return}{Return}
\caption{CALM}
\Input{LLM $\pi_\theta$, Evaluation environment $g$, number $G$ of responses to be sampled for one prompt, maximum round number $T$, Population size $L_\mathrm{p}$, Sampling weight $\bm{w}$ for each operator, Hyperparameter $\delta_0$ and $C$ that control the collapse mechanism, set of seed heuristic $\mathcal{H}_\mathrm{seed}$ (set to be $\emptyset$ if not given any seed heuristic).}
Initialize collapse counter $t_c=-1$, best heuristic $h^*=\mathrm{null}$, best performance $g^*=-\infty$, heuristic pool $\mathcal{H}_\mathrm{pool}=\mathcal{H}_\mathrm{seed}$, $w_{i}=\bm{w}_{\mathrm{injection}}$\;
\For{$t=1,\cdots, T$}{
Operator base $\mathrm{OPs}\leftarrow\{\mathrm{Initialization}\}$\;
\If{$\lvert\mathcal{H}_\mathrm{pool}\rvert\geq1$}{
$\mathrm{OPs}\leftarrow\{\mathrm{Injection}, \mathrm{Replacement}, \mathrm{Crossover},\mathrm{Simplification}\}$\;
}
\If{$\lvert\mathcal{H}_\mathrm{pool}\rvert\geq2$}{
$\mathrm{OPs}\leftarrow\mathrm{OPs}\cup\{\mathrm{Crossover}\}$\;
}
\uIf{$\lvert\mathcal{H}_\mathrm{pool}\rvert<L_\mathrm{p}$}{
$\bm{w}_\mathrm{Injection}\leftarrow\max(\bm{w})$\;
}
\uElse{
$\bm{w}_\mathrm{Injection}\leftarrow w_{i}$\;
}
$\mathcal{H}_\mathrm{base}\leftarrow\emptyset$, $\mathrm{op}\leftarrow$ Draw an operator from $\mathrm{OPs}$ with the probability proportional to $\bm{w}$\;
\If{$\mathrm{op}\neq\mathrm{Initialization}$}{
$h_{c, 1}\leftarrow$Draw an heuristic from top-$L_\mathrm{p}$-performing heuristics in $\mathcal{H}_\mathrm{pool}$, where the sampling probability of an heuristic $h$ is proportional to $1/\mathrm{rank}_p(h)$ and $\mathrm{rank}_p(h)$ is the heuristic's performance rank\;
$\mathcal{H}_\mathrm{base}\leftarrow\mathcal{H}_\mathrm{base}\cup\{h\}$\;
\If{$\mathrm{op}=\mathrm{Crossover}$}{
    \uIf{$\mathrm{random}(0, 1)\leq0.5$}{
        $h_{c,2}\leftarrow$Draw a heuristic from the population by performance rank as sampling $h_{c,1}$\;
    }
    \uElse{
        Calculate diversity metric $\mathrm{div}(h_{c,1}, h)=\frac{\lvert\mathrm{idea\_token}(h)\setminus\mathrm{idea\_token}(h_{c,1})\rvert}{\lvert\mathrm{idea\_token}(h)\rvert},\forall h\in\mathcal{H}_\mathrm{pool}$\;
        $h_{c,2}\leftarrow$Draw a heuristic from the pool by diversity rank where the sampling probability is proportional to $1/\mathrm{rank}_d(h)$ (a larger diversity value yields a higher probability)\;
    }
    $\mathcal{H}_\mathrm{base}\leftarrow\mathcal{H}_\mathrm{base}\cup\{h_{c,2}\}$\;
}
}
$q\leftarrow$Generate prompt by the operator $\mathrm{op}$ and base heuristics $\mathcal{H}_\mathrm{base}$\;
$\mathcal{O}\leftarrow$Sample $G$ responses from $\pi_\theta$ for $q$\;
$\mathcal{H}_\mathrm{feasible}, \hat{r}_\mathcal{O}\leftarrow$ Try extracting a feasible heuristic from each response $o\in\mathcal{O}$ and assign reward to each response following Section~\ref{sec:reward}\;
$\theta\leftarrow$Update the LLM by GRPO that optimizes Equation~\ref{eq:grpo_obj} with $(q, \mathcal{O}, \hat{r}_\mathcal{O})$\;
$\mathcal{H}_\mathrm{pool}\leftarrow\mathcal{H}_\mathrm{pool}\cup\mathcal{H}_\mathcal{feasible}$\;
$h^*\leftarrow\arg\max_{h\in\mathcal{H}_\mathrm{pool}}g(h)$\;
\uIf{$g(h^*)=g^*$ and $\lvert\mathcal{H}_\mathrm{pool}\rvert\geq L_\mathrm{p}$}{
    \tcc{If the population is full, the counter for collapse starts.}
    $t_c\leftarrow \max({t_c, 0}) + 1$\;
}\uElse{
    $g^*=(h^*), t_c\leftarrow\min(t_c,0)$\;
}
\If{$\mathrm{random(0,1)}\leq \delta_0t_c$ or $t_c\geq C$}{
    $\mathcal{H}_\mathrm{base}\leftarrow\{h^*\}\cup\mathcal{H}_\mathrm{seed},t_c\leftarrow-1$;  \tcc{Collapse}
}
}
\Return{$h^*$}
\end{algorithm}

\section{Prompts Used in CALM}\label{appendix-sec:prompts}

\begin{figure}
    \centering
    \includegraphics[width=\linewidth]{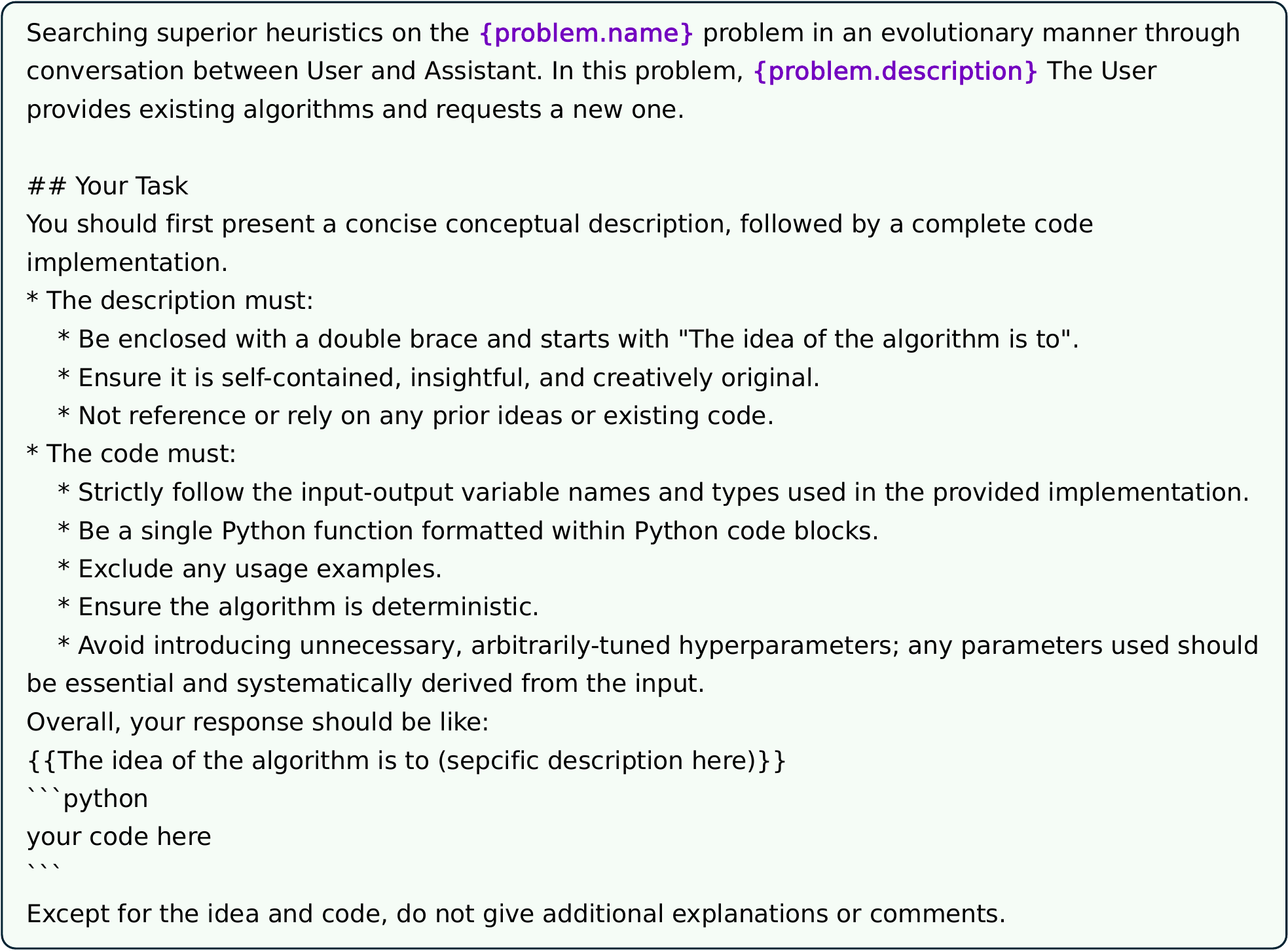}
    \caption{Template of the system prompt.}
    \label{fig:system-prompt-template}
\end{figure}

\textbf{System Prompt.}
The system prompt is generated by inserting the name and description into the template shown in Figure~\ref{fig:system-prompt-template}. The specific prompt used for each problem can be found in Table~\ref{tab:prompt-for-each-problem}.

\begin{figure}
    \centering
    \includegraphics[width=\linewidth]{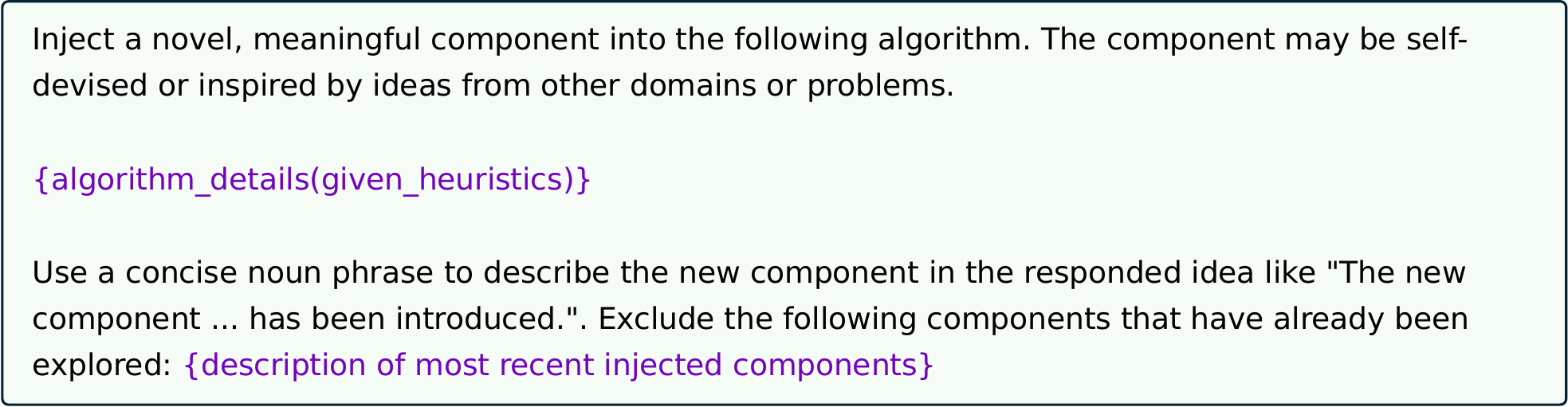}
    \caption{Template of the injection prompt.}
    \label{fig:injection-prompt-template}
\end{figure}

\begin{figure}
    \centering
    \includegraphics[width=\linewidth]{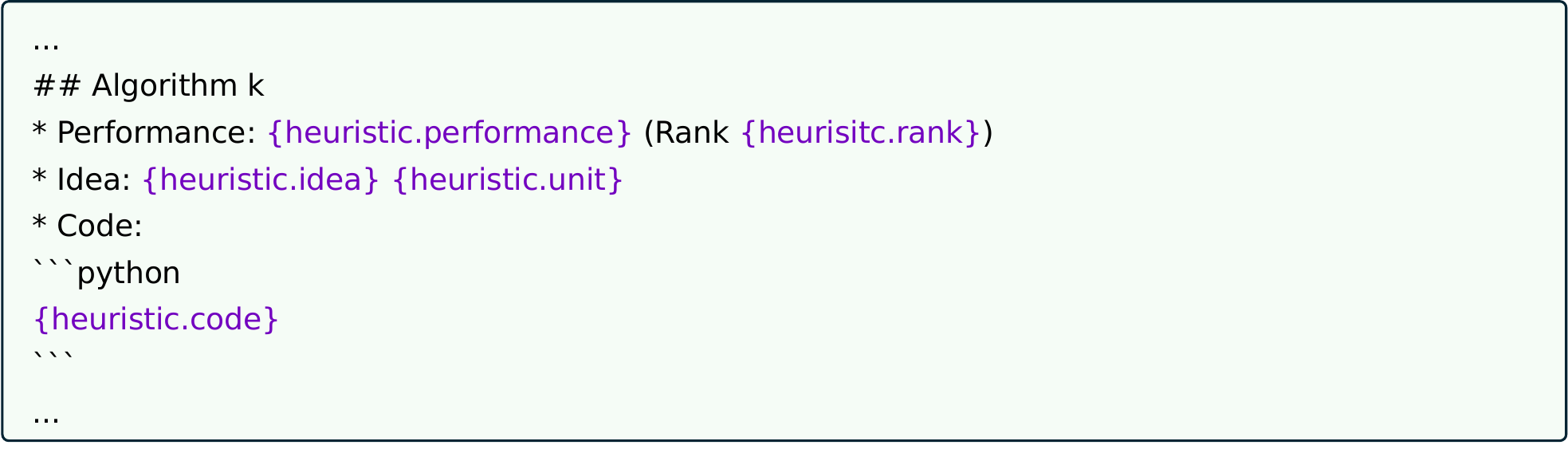}
    \caption{Template of algorithm details.}
    \label{fig:algo-details-prompt-template}
\end{figure}

\textbf{Injection Prompt.}
The template used to generate injection prompts is shown in Figure~\ref{fig:injection-prompt-template}. In the prompt template, the algorithm details are generated by the given heuristics and the prompt template in Figure~\ref{fig:algo-details-prompt-template}. The description of the most recent injected components is created by (1) parsing the string wrapped within "The new component ... has been introduced", (2) globally saving the historical new components, and (3) picking the last 10 new components to be used.

\begin{figure}
    \centering
    \includegraphics[width=\linewidth]{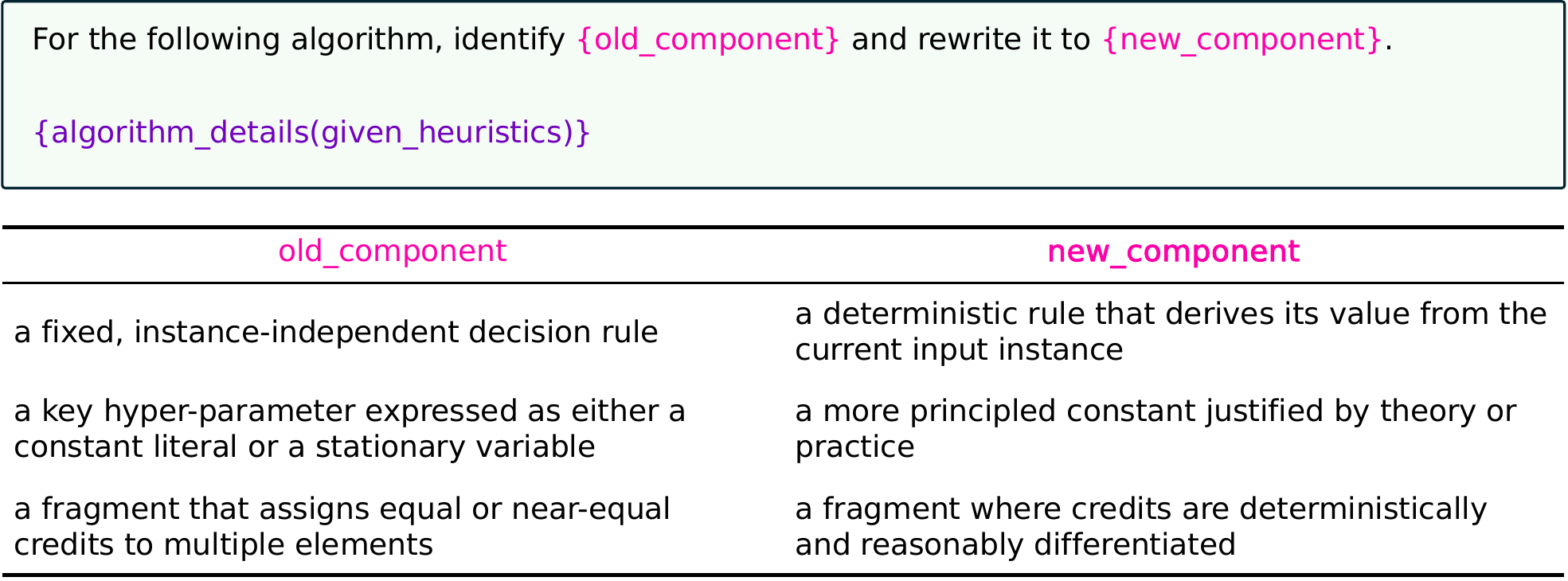}
    \caption{Template of the replacement prompt.}
    \label{fig:replacement-prompt-template}
\end{figure}

\textbf{Replacement Prompt.}
The replacement prompt is created by the template, some predefined component Paris shown in Figure~\ref{fig:replacement-prompt-template}, and the algorithm detail template shown in Figure~\ref{fig:algo-details-prompt-template}.

\begin{figure}
    \centering
    \includegraphics[width=\linewidth]{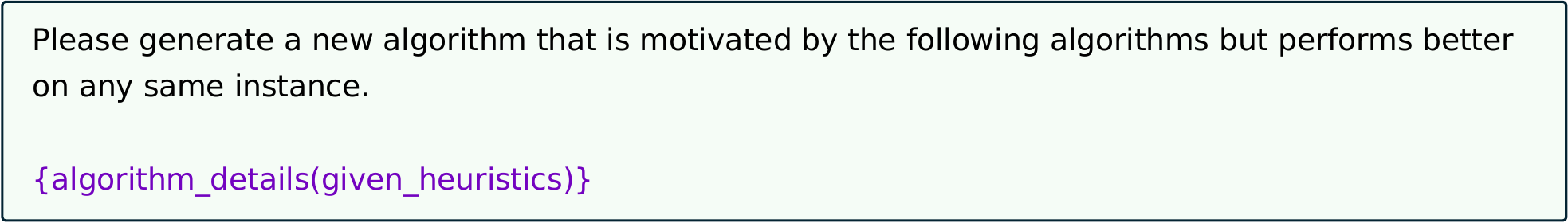}
    \caption{Template of the crossover prompt.}
    \label{fig:crossover-prompt-template}
\end{figure}

\textbf{Crossover Prompt.}
The crossover prompt is generated by the template shown in Figure~\ref{fig:crossover-prompt-template} and the algorithm detail template shown in Figure~\ref{fig:algo-details-prompt-template}.

\begin{figure}
    \centering
    \includegraphics[width=\linewidth]{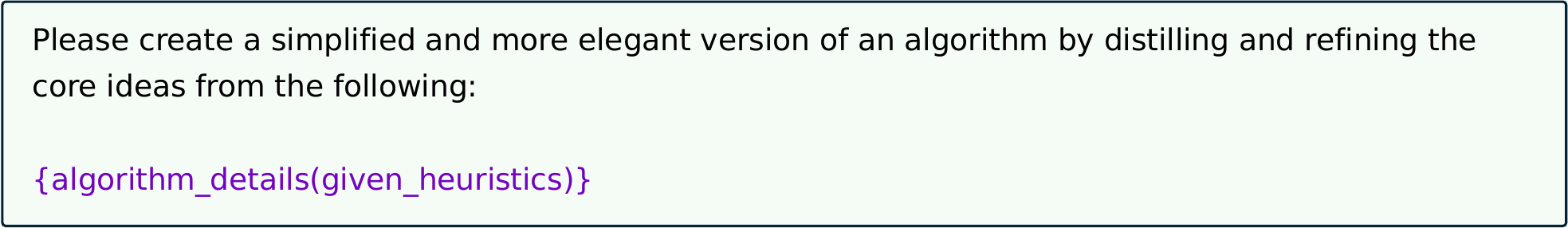}
    \caption{Template of the simplification prompt.}
    \label{fig:simplification-prompt-template}
\end{figure}

\textbf{Simplification Prompt.}
The simplification prompt is created by the template shown in Figure~\ref{fig:simplification-prompt-template}
 and the algorithm detail template shown in Figure~\ref{fig:algo-details-prompt-template}.

\begin{figure}
    \centering
    \includegraphics[width=\linewidth]{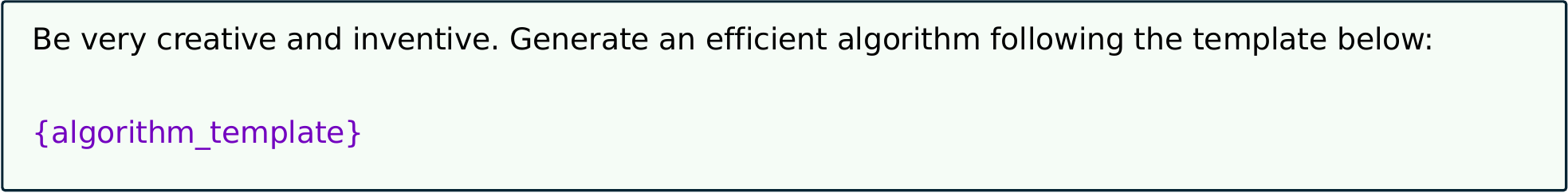}
    \caption{Template of the initialization prompt.}
    \label{fig:initialization-prompt-template}
\end{figure}

\textbf{Initialization Prompt.}
The initialization prompt is created by the template shown in Figure~\ref{fig:initialization-prompt-template}. The algorithm template is a function signature. 

\begin{table}[htbp]
  \centering
  \caption{Information of each problem used in prompts}
  \label{tab:prompt-for-each-problem}

  \begin{tabular*}{\linewidth}{@{\extracolsep{\fill}}
      P{0.10}   
      P{0.155}  
      P{0.255}  
      P{0.225}  
      P{0.145}  
    }
    \toprule
       & \textbf{OBP}
       & \textbf{CVRP}
       & \textbf{OP}
       & \textbf{TSP} \\
    \midrule
    Name
       & Online Bin Packing
       & Capacitated Vehicle Routing
       & Orienteering
       & Traveling Salesman \\
    Unit
       & percent of the gap to the lower bound
       & units of travel distance
       & units of collected reward
       & length units of the tour \\
    Description
       & Items arrive sequentially and must be placed immediately into bins only if they fit within the remaining capacity. The objective is to minimize the number of bins used.
       & A fleet of vehicles with limited carrying capacity must deliver goods to a set of geographically distributed customers with known demands, minimizing the total travel distance while ensuring that no vehicle exceeds its capacity. The goal is to design a heuristic function that estimates the desirability of moving between customers, to be used within an Ant Colony Optimization (ACO) algorithm.
       & An agent must visit a subset of locations, each offering a reward, within a maximum travel budget. The objective is to maximize the total collected reward while adhering to the travel constraint. The goal is to design a heuristic function that estimates the desirability of moving between locations, to be used within an Ant Colony Optimization (ACO) algorithm.
       & The algorithm must find a tour that visits each node exactly once and returns to the start node. The objective is to minimize the length of the tour. \\
    \bottomrule
  \end{tabular*}
\end{table}

\clearpage
\section{Discussion about the Diversity-Aware Crossover Operator}\label{appendix-sec:crossover}
Notably,\cite{mcts-ahd} allowed heuristic selection beyond the top-performing population, offering greater exploration flexibility, though without explicitly modeling diversity. In contrast,\cite{hsevo} emphasized the role of diversity in heuristic evolution but did not integrate it into crossover and operated within a fixed-size population. Therefore, CALM’s crossover operator complements prior work by explicitly incorporating diversity into the crossover process.

\section{More Details For the Collapse Mechanism}\label{appendix-sec:collapse}

\subsection{Proof of Equation (\ref{eq:collapse-round-expectation})}

Let \(c_n\) be the stagnation counter just before collapse.  Under the collapse mechanism with per‐round hazard
\[
p_k = k\,\delta_0,\qquad k=1,2,\dots,
\]
the probability of surviving beyond \(k\) rounds is
\[
\Pr[c_n>k] \;=\;\prod_{i=1}^{k}\bigl(1 - i\,\delta_0\bigr),
\]
which vanishes for \(k\ge\lfloor1/\delta_0\rfloor\).

By definition,
\[
\mathbb{E}[\,c_n\,]
=\sum_{k=1}^\infty k\,\Pr[c_n=k].
\]
Introduce the nonnegative array
\[
a_{j,k} =
\begin{cases}
\Pr[c_n=k], & k\ge j\ge1,\\
0, & \text{otherwise}.
\end{cases}
\]
Then
\begin{align*}
\sum_{k=1}^\infty k\,\Pr[c_n=k]
&= \sum_{k=1}^\infty\sum_{j=1}^k \Pr[c_n=k]
= \sum_{k=1}^\infty\sum_{j=1}^\infty a_{j,k}.
\end{align*}
Since \(a_{j,k}\ge0\), Tonelli’s theorem allows swapping the sums:
\[
\sum_{k=1}^\infty\sum_{j=1}^\infty a_{j,k}
= \sum_{j=1}^\infty\sum_{k=1}^\infty a_{j,k}
= \sum_{j=1}^\infty\sum_{k=j}^\infty \Pr[c_n=k]
= \sum_{j=1}^\infty \Pr[c_n\ge j]
= \sum_{j=0}^\infty \Pr[c_n>j].
\]
Hence the tail‐sum identity
\[
\mathbb{E}[\,c_n\,]
= \sum_{j=0}^\infty \Pr[c_n>j].
\]

For \(\delta_0\ll1\) we approximate the finite product by exponentiating its logarithm, using the Maclaurin expansion
\[
\ln(1-x)
= -\sum_{m=1}^\infty \frac{x^m}{m}
= -x-\tfrac{x^2}{2}-\cdots,
\quad |x|<1,
\]
with \(x=i\delta_0\).  Truncating at the linear term gives
\[
\sum_{i=1}^k \ln\bigl(1 - i\,\delta_0\bigr)
\approx -\sum_{i=1}^k i\,\delta_0
= -\tfrac{\delta_0}{2}\,k(k+1)
\approx -\tfrac{\delta_0}{2}\,k^2,
\]
so
\[
\Pr[c_n>k]
\approx \exp\!\bigl(-\tfrac{\delta_0}{2}k^2\bigr).
\]
Substituting into the tail‐sum and replacing the discrete sum by an integral yields
\[
\mathbb{E}[\,c_n\,]
\approx \sum_{k=0}^\infty e^{-\frac{\delta_0}{2}k^2}
\approx \int_0^\infty e^{-\frac{\delta_0}{2}x^2}\,dx
= \sqrt{\frac{\pi}{2\,\delta_0}},
\]
which establishes Equation (\ref{eq:collapse-round-expectation}).

\subsection{Goodness}
Following this reset, the search effectively starts anew, but with a strategic advantage: it builds upon the best insights discovered so far. Importantly, during the early stage of repopulation, the system temporarily relaxes selection constraints. New heuristics generated via injection, replacement, or crossover are allowed into the population regardless of performance, as long as the total number of heuristics remains below the target population size. This gives structurally novel but potentially suboptimal components the opportunity to propagate and evolve—something not feasible under normal selection pressure, where only top-performing heuristics are retained and processed further.

\section{Justification for Penalizing Random Modules}
Randomized heuristics are excluded in this work because their stochastic behavior substantially increases evaluation cost and noise. To enforce determinism, CALM penalizes responses that invoke randomness (e.g., usage of \texttt{random}, \texttt{np.random}, etc.). The framework could support randomized heuristics by relaxing this constraint, though evaluation overhead would increase.

\section{More Experimental Details}\label{appendix-sec:running-time}
\subsection{Implementation Details}
We build CALM on Unsloth~\cite{unsloth}, with two modifications: raising the learning rate to $5\times10^{-5}$ for faster adaptation and sampling $G=4$ responses per prompt to enable more evolutionary steps under a fixed query budget.

We set the initial collapse growth rate to $\delta_0 = 0.0005$ (max threshold $C=25$), cap training at $T = 500$ rounds, and assign operator sampling probabilities in the ratio $1:1:2:4$ for simplification, injection, modification, and crossover, respectively. Each heuristic is evaluated within 60 s~\cite{mcts-ahd}. All experiments ran on a 24 GB NVIDIA A30 GPU with an Intel Xeon Gold 5220R CPU.

Under this configuration, the average running time of CALM for the OBP, CVRP, OP, and TSP is about 6.8, 7.2, 5.3, and 5.5 hours, respectively, for $T=500$ steps. However, it is important to note that the actual running time for a single trial may vary considerably due to the stochastic nature of the LLM and the potentially large number of heuristics generated, each requiring time-intensive evaluation.

\subsection{Justification For the Baseline Implementations}
The source code, training dataset, and test dataset for AlphaEvolve~\cite{alphaevolve} and EvoTune~\cite{evotune} are not available. Besides, ReEvo~\cite{reevo} and its follow-up approach HSEvo~\cite{hsevo} can stop at a very early stage in evolution as found by~\cite{mcts-ahd}. Thus, the results of them on TSP are not reported.

\subsection{Description of Problems in Experiments}\label{appendix-sec:problem}

\textbf{Online Bin Packing (OBP).}
A sequence of items of varying sizes arrives one by one. Each bin has a fixed capacity. Upon arrival of an item, the algorithm must immediately assign it to an existing bin that has enough remaining space or open a new bin. The goal is to minimize the total number of bins used. The input of the heuristic is the size of the current item and the remaining capacities of the bins. The output of the heuristic is the priority score of each observed bin, where the feasible bin with the highest score will be selected to accomodate the item.

\textbf{Traveling Salesman Problem (TSP) under Step-by-Step Construction.}
Given a set of locations with pairwise travel distances, the objective is to construct a tour that starts at one location, visits each other location exactly once, and returns to the start. At each step the heuristic must choose the next unvisited location based solely on the information gathered so far. The aim is to keep the total travel distance as small as possible. 

\textbf{Capacitated Vehicle Routing Problem (CVRP) under ACO.}
A fleet of vehicles with identical load capacity must serve a set of customers, each with a known demand, and all vehicles start and end at a central depot. Under the Ant Colony Optimization framework, many artificial “ants” build routes by moving from customer to customer. Each ant’s choice of next customer is guided by a combination of pheromone trails—updated based on previous high-quality solutions—and heuristic scores provided by the LLM. The goal is to serve all customers while minimizing the total distance traveled and respecting vehicle capacity limits.

\textbf{Orienteering Problem (OP) under ACO.}
Starting from a given location (and possibly ending at the same or another specified location), an agent may visit a subset of available sites, each offering a reward, subject to an overall travel budget. Within the ACO framework, ants construct candidate paths by choosing which site to visit next based on pheromone levels and LLM-generated heuristic scores that estimate the benefit of each edge under the reward-and-budget trade-off. The aim is to collect as much reward as possible without exceeding the travel budget.

\subsection{Generated Heurisitcs}
\definecolor{keywordcolor}{RGB}{0,0,180}
\definecolor{commentcolor}{RGB}{63,127,95}
\definecolor{stringcolor}{RGB}{163,21,21}
\definecolor{bgcolor}{RGB}{248,248,248}

\lstdefinestyle{pythonstyle}{
  language=Python,
  basicstyle=\ttfamily\small,
  keywordstyle=\color{keywordcolor}\bfseries,
  commentstyle={\ttfamily\small\color{commentcolor}},
  stringstyle=\color{stringcolor},
  backgroundcolor=\color{bgcolor},
  frame=single,
  framesep=5pt,
  rulecolor=\color{gray!50},
  numbers=none,            
  showstringspaces=false,
  tabsize=4,
  breaklines=true,
  breakatwhitespace=true,
}

\lstnewenvironment{heuristic}[2][]{
  \lstset{
    style=pythonstyle,
    caption={#2},
    captionpos=b,
    #1
  }
}{}
\renewcommand{\lstlistingname}{Heuristic}

\begin{heuristic}{OBP, by CALM (local, w/ GRPO)}
"""
The idea of the algorithm is to refine the scoring mechanism by introducing logarithmic adjustments and a novel scoring component that captures the logarithmic relationship between the remaining capacity and the square of the item size, and an adjusted logarithmic density term that provides a more refined scoring mechanism. This new algorithm aims to enhance the accuracy of bin suitability assessment by adding a component that adjusts the score based on the logarithmic difference between the remaining capacity and the maximum bin capacity. The algorithm also simplifies the scoring steps to make it more elegant and efficient.
"""

import numpy as np

def step(item_size: float, remaining_capacity: np.ndarray) -> np.ndarray:
    max_bin_cap = max(remaining_capacity)
    bin_density = np.sum(remaining_capacity) / (item_size * len(remaining_capacity))
    log_adj = np.log(remaining_capacity + 1) / np.log(max_bin_cap + 1)
    score = (remaining_capacity - max_bin_cap)**2 / item_size + remaining_capacity**2 / (item_size**2) + remaining_capacity**2 / (item_size**3) + bin_density * remaining_capacity
    
    score[remaining_capacity > item_size] = -score[remaining_capacity > item_size]
    score[1:] -= score[:-1]
    
    score *= log_adj
    score += log_adj * remaining_capacity
    
    score *= log_adj
    new_component = remaining_capacity / (item_size - remaining_capacity + 1)
    score += new_component
    
    new_component = remaining_capacity * np.log(remaining_capacity + 1) / (item_size * np.log(max_bin_cap + 1)) * (1 - remaining_capacity / item_size)
    score += new_component
    
    new_adjustment = (remaining_capacity / item_size) * log_adj
    score += new_adjustment
    
    remaining_capacity_adjusted = remaining_capacity / item_size
    score += np.log(remaining_capacity_adjusted + 1) / np.log(max_bin_cap + 1)
    
    new_component = (remaining_capacity - 1) / (item_size - remaining_capacity + 1) * log_adj / np.log(max_bin_cap + 1)
    score += new_component
    
    new_component = log_adj * remaining_capacity / (item_size - remaining_capacity)
    score += new_component
    
    new_component = remaining_capacity * np.log(remaining_capacity + 1) / (item_size**2) * (1 - remaining_capacity / item_size)
    score += new_component
    
    return score
\end{heuristic}

\begin{heuristic}{OBP, by CALM (API, w/o GRPO)}
"""
The idea of the algorithm is to introduce the "Bin Utilization Diminution" component, which assesses the degree of bin usage throughout the sequence of placements and introduces a diminishing incentive for overpopulating any particular bin beyond a certain threshold. This encourages a more even distribution of item placements across all bins, thereby reducing the risk of reaching capacity too quickly in any single bin, helping to extend the lifespan and utility of each bin in the packing process. By dynamically adjusting the fit score to favor items that contribute to a balanced utilization, the algorithm aims to enhance overall bin efficiency and minimize the total bin count.
"""

import numpy as np

def step(item_size: float, remaining_capacity: np.ndarray) -> np.ndarray:
    avg_item_size = np.mean(item_size) if item_size > 0 else 1.0
    adaptive_factor = avg_item_size / np.maximum(remaining_capacity, 1e-10)

    fit_score = np.maximum(remaining_capacity - item_size, 0) / (remaining_capacity + 1e-10)
    fit_score[remaining_capacity < item_size] = -np.inf

    sustainability_score = (remaining_capacity - avg_item_size) ** 2
    sustainability_score[remaining_capacity < item_size] = np.inf

    historical_fit_scores = np.cumsum(fit_score)
    normalized_historical_fit_scores = historical_fit_scores / (np.max(historical_fit_scores) + 1e-10)

    combined_scores = (
        (0.5 * fit_score * adaptive_factor) + 
        (0.3 / (sustainability_score + 1e-10)) - 
        (0.2 * normalized_historical_fit_scores)
    )

    differentiation_factor = 1 / (1 + np.arange(len(remaining_capacity)) * 0.1)
    combined_scores *= differentiation_factor

    cumulative_fit_impact = np.cumsum(fit_score) / (np.arange(1, len(remaining_capacity) + 1) + 1)
    cumulative_fit_adjustment = np.maximum(fit_score - cumulative_fit_impact, 0)

    combined_scores += 0.4 * cumulative_fit_adjustment

    temporal_utilization_metric = np.arange(len(remaining_capacity)) / (np.maximum(remaining_capacity, 1e-10) + 1e-10)
    combined_scores *= (1 + temporal_utilization_metric)

    sequential_elasticity = np.exp(-np.arange(len(remaining_capacity)) / (np.mean(np.maximum(remaining_capacity, 1e-10)) + 1e-10))
    combined_scores *= sequential_elasticity

    size_factor = 1 + (item_size / (np.sum(item_size) + 1e-10))

    # New Component: Bin Utilization Diminution
    overutilization_penalty = np.maximum(0, np.cumsum(item_size) / (np.maximum(np.cumsum(remaining_capacity), 1e-10) + 1e-10) - 1)
    combined_scores -= 0.3 * overutilization_penalty  # Encourage even distribution across bins

    # Eventual Capacity Influence
    eventual_capacity_score = np.log(np.maximum(np.arange(1, len(remaining_capacity) + 1), 1)) / (np.maximum(remaining_capacity, 1e-10) + 1e-10)
    combined_scores -= 0.3 * eventual_capacity_score  # Penalize bins that don't contribute to optimal utilization

    distinct_scores = combined_scores * size_factor

    return distinct_scores
\end{heuristic}

\begin{heuristic}{CVRP, by CALM(local, w/ GRPO)}
"""
The idea of the algorithm is to further refine the savings potential calculation by emphasizing a more adaptive balance factor that is influenced by the current instance's capacity utilization and the diversity of capacity usage across the routing problem. By leveraging a more sophisticated adaptive balance factor and reducing the complexity of the penalty factor, we ensure that nodes that are too close to each other are penalized appropriately without overly compounding the impact. This simplified yet adaptive approach allows for a nuanced exploration of the solution space, enhancing the ACO algorithm's ability to converge to high-quality solutions while maintaining a balance between exploration and exploitation. Additionally, we introduce a clustering-based adjustment factor that captures the overall network connectivity and adjusts the savings potential accordingly, leading to more robust and flexible routing plans.
"""

import numpy as np

def advanced_heuristics_v7(distance_matrix: np.ndarray, coordinates: np.ndarray, demands: np.ndarray, capacity: int) -> np.ndarray:
    capacity_prob = demands / capacity
    distance_reciprocal = 1 / distance_matrix
    proximity_factor = np.linalg.norm(coordinates[:, np.newaxis, :] - coordinates[np.newaxis, :, :], axis=2)
    proximity_factor /= np.max(proximity_factor)  # Normalize between 0 and 1
    proximity_factor = 1 - proximity_factor  # Invert for higher penalty as proximity increases
    
    remaining_demands = capacity - demands
    future_savings = (remaining_demands[:, np.newaxis] * remaining_demands) / (distance_matrix * (remaining_demands[:, np.newaxis] + remaining_demands))
    capacity_ratio = remaining_demands / capacity
    proximity_savings = proximity_factor * capacity_ratio
    
    # Cluster-based proximity adaptive savings potential
    cluster_savings = np.zeros_like(distance_matrix)
    cluster_distance = np.sum(distance_matrix, axis=1) / np.linalg.norm(capacity_prob - 1, ord=1)
    cluster_adj_factor = (remaining_demands[:, np.newaxis] * remaining_demands * cluster_distance ** 3.5) / (distance_matrix * (remaining_demands[:, np.newaxis] + remaining_demands))
    
    # Adaptive balance factor adjusted based on remaining capacity and cluster adjustment
    balance_factor = np.min([1, 0.975 + 0.05 * capacity_prob.mean() + 0.03 * cluster_adj_factor.mean() + 0.005 * np.var(capacity_prob)])
    
    # Penalty factor that heavily penalizes nodes that are too close to each other, focusing on the proximity to the next node
    penalty_factor = proximity_factor ** 3
    
    # Combine all components
    probability = distance_reciprocal * capacity_prob * proximity_factor * future_savings * proximity_savings * cluster_adj_factor * (1 - balance_factor + proximity_savings * balance_factor) * (1 - penalty_factor) * (1 + cluster_adj_factor * 0.6)
    return probability
\end{heuristic}

\begin{heuristic}{CVRP, by CALM (API, w/o GRPO)}
"""
The idea of the algorithm is to refine the credit allocation process in the vehicle routing problem by implementing a deterministic weighting mechanism that assigns distinct credits to customers based on their delivery demands, individual distance factors, and their influence on overall routing efficiency, thus ensuring that credits reflect meaningful differences without redundancy.
"""

import numpy as np
from sklearn.cluster import DBSCAN

def heuristics(distance_matrix: np.ndarray, coordinates: np.ndarray, demands: np.ndarray, capacity: int) -> np.ndarray:
    num_customers = demands.shape[0]
    cumulative_penalty = np.zeros(num_customers)

    # Calculate baseline scores from demand to distance with added urgency weighting
    urgency_weight = np.linspace(1, 1.5, num_customers)
    base_score = (demands * urgency_weight) / (distance_matrix + 1e-5)
    base_score[np.isnan(base_score)] = 0

    # Set penalties for exceeding capacity based on cumulative demands
    for i in range(num_customers):
        current_demand = demands[i]
        cumulative_penalty[i] = max(0, current_demand - capacity)

    # Normalize distances to emphasize closer customers to refine scoring
    normalized_distance_score = 1 / (np.clip(distance_matrix, 1e-5, None) ** 2.5)

    # Calculate effective capacity utilization adjustment
    effective_capacity_utilization = np.clip((capacity - demands) / capacity, 0, 1)

    # Historical performance adjustments
    historical_performance_factor = np.zeros(num_customers)
    for i in range(num_customers):
        historical_performance_factor[i] = np.mean([base_score[j] for j in range(num_customers) if distance_matrix[i][j] < 10 and j != i])

    # Spatial clustering mechanism
    clustering_model = DBSCAN(eps=5, min_samples=2).fit(coordinates)
    labels = clustering_model.labels_
    cluster_scores = np.zeros(num_customers)

    # Calculate cluster-based scores with deterministic differentiation
    for cluster_id in set(labels):
        if cluster_id != -1:  # Ignore noise points
            cluster_indices = np.where(labels == cluster_id)[0]
            total_demand = demands[cluster_indices].sum()
            for idx in cluster_indices:
                # Implement differentiated scoring based on demand, ensuring non-equal credits
                cluster_demand_factor = (demands[idx] / total_demand) if total_demand > 0 else 0
                distance_weight = 1 / (1 + distance_matrix[idx].min())  # Closer customers get more weight
                cluster_scores[idx] = cluster_demand_factor * distance_weight  # Mix demand and distance

    # New resilience score based on historical demand variability
    demand_variability = np.std(demands)
    resilience_score = 1 / (1 + demand_variability)

    # Compose final scores combining all elements including the new resilience score
    final_scores = base_score * normalized_distance_score * effective_capacity_utilization * (1 + historical_performance_factor + cluster_scores) * resilience_score

    return final_scores
\end{heuristic}

\begin{heuristic}{OP, by CALM (local, w/ GRPO)}
"""
The idea of the algorithm is to refine the exploration-expemy exploitation trade-off by introducing a sinusoidal decay that incorporates a sinusoidal penalty with a sinusoidal smoothness adjustment. This adjustment helps to smooth the preference for both recent and distant nodes, leading to a more balanced and improved performance.
"""

import numpy as np

def enhanced_heuristics(prize: np.ndarray, distance: np.ndarray, maxlen: float) -> np.ndarray:
    # Exponential decay for immediate high Subscription nodes
    exp_ratio = np.exp(prize[np.newaxis, :] / distance - maxlen)
    
    # Logarithmic scaling for exploration
    log_ratio = np.log(prize[np.newaxis, :] + 1) / distance
    
    # Sinusoidal decay for recent nodes with a sinusoidal smoothness adjustment
    sinusoidal_penalty = 0.5 * (1 + np.sin(np.pi * distance / (maxlen + 1))) * (distance / maxlen) * maxlen
    
    # Combined ratio
    combined_ratio = exp_ratio * log_ratio * (1 - sinusoidal_penalty)
    
    # Ensure the ratio is non-negative
    combined_ratio[combined_ratio < 0] = 0
    
    return combined_ratio
\end{heuristic}

\begin{heuristic}{OP, by CALM (API, w/o GRPO)}
"""
The idea of the algorithm is to introduce a novel component called "reward fluctuation sensitivity" which adjusts the desirability of each location based on the variability of rewards over time. This component accounts for the possibility that rewards may change or fluctuate due to external factors, thereby allowing the agent to prioritize locations not only by their current rewards but also by the potential volatility of those rewards. This sensitivity is integrated into the existing framework, allowing for a more dynamic response to the changing landscape of rewards, ultimately enhancing the agent\'s decision-making process and route optimization.
"""

import numpy as np

def heuristics(prize: np.ndarray, distance: np.ndarray, maxlen: float) -> np.ndarray:
    adjusted_distance = distance + 1e-10  # Avoid division by zero
    potential_reward = np.zeros_like(prize)

    for i in range(len(prize)):
        reachable_indices = np.where(distance[i] <= maxlen)[0]
        potential_reward[i] = np.sum(prize[reachable_indices]) if reachable_indices.size > 0 else 0

    reward_hist_factor = potential_reward / (1 + np.sum(prize))  # Shape reward based on historical performance
    reward_decay = np.exp(-adjusted_distance / maxlen)  # Decay effect for distant rewards

    proximity_factor = (maxlen - adjusted_distance) ** 4  # Further enhance proximity impact with quartic distance
    proximity_factor[proximity_factor < 0] = 0

    tiered_adjustment = (prize / (adjusted_distance + 1e-10)) ** 2  # Classify rewards into categories for tiering

    # Reward volatility assessment component
    volatility_factor = np.zeros_like(prize)
    for i in range(len(prize)):
        historical_rewards = prize[np.where(distance[i] <= maxlen)[0]]
        if historical_rewards.size > 1:
            volatility_factor[i] = np.std(historical_rewards) / np.mean(historical_rewards)  # Coefficient of variation

    # Risk-reward analysis component
    variability_factor = np.zeros_like(prize)
    for i in range(len(prize)):
        historical_rewards = prize[np.where(distance[i] <= maxlen)[0]]
        if historical_rewards.size:
            variability_factor[i] = np.mean(historical_rewards) - np.std(historical_rewards)  # Basic differentiation 

    final_heuristic = (reward_hist_factor * reward_decay * proximity_factor *
                       tiered_adjustment) / (1 + volatility_factor + variability_factor + 1e-10)  
    return final_heuristic
\end{heuristic}

\begin{heuristic}{TSP, by CALM (local, w/ GRPO)}
"""
The idea of the algorithm is to select the next node by optimizing a heuristic that considers the distance to the current node, the average distance to unvisited nodes, the variance of distances to the current node from the unvisited nodes, the entropy of distances to the destination node from each of the unvisited nodes, the average distance from the destination node to each of the unvisited nodes, the current node's distance to the destination node, and the standard deviation of the overall tour distances. This proposed algorithm aims to introduce a new term that captures the deviation of the current node from the average tour length and balances it with the entropy term to reduce the overall tour length. Additionally, this method assigns more weight to the standard deviation of the distances from the destination node to each of the unvisited nodes, which helps in reducing the variability of distances and thus leading to more consistent and shorter tour lengths.
"""

import numpy as np

def select_next_node(current_node: int, destination_node: int, unvisited_nodes: set, distance_matrix: np.ndarray) -> int:
    scores = {}

    for node in unvisited_nodes:
        all_distances = [distance_matrix[node][i] for i in unvisited_nodes if i != node]
        average_distance = np.mean(all_distances)
        standard_deviation = np.std(all_distances)
        variance_of_distances = np.var([distance_matrix[current_node][i] for i in unvisited_nodes if i != node])
        entropy_of_distances = -np.sum(np.log2([distance_matrix[destination_node][i] for i in unvisited_nodes if i != node]) / len(unvisited_nodes))
        average_distance_to_destination = np.mean([distance_matrix[destination_node][i] for i in unvisited_nodes if i != node])

        score = (
            0.6 * distance_matrix[current_node][node]
            - 0.4 * average_distance
            + 0.3 * standard_deviation
            - 0.2 * entropy_of_distances
            - 0.1 * distance_matrix[destination_node][node]
            - 0.08 * variance_of_distances
            - 0.05 * average_distance_to_destination
            - 0.01 * (np.mean([distance_matrix[current_node][i] for i in unvisited_nodes]) - average_distance)
            - 0.005 * entropy_of_distances
            - 0.008 * distance_matrix[current_node][node] * distance_matrix[node][destination_node]
            - 0.006 * standard_deviation * distance_matrix[node][destination_node]
        )
        scores[node] = score

    next_node = min(scores, key=scores.get)
    return next_node
\end{heuristic}

\begin{heuristic}{TSP, by CALM (API, w/o GRPO)}
"""
The idea of the algorithm is to select the next node to visit from the unvisited nodes, incorporating a novel component of dynamic path optimization feedback. The new component analyzes previous decision points in the tour to determine the effectiveness of the routes taken, adjusting future node selection to favor pathways that have historically resulted in lower overall traversal costs. This method not only enhances the algorithm's ability to learn from its own experiences but also promotes the selection of routes that align with optimal connectivity patterns established during the tour.
"""

import numpy as np

def select_next_node(current_node: int, destination_node: int, unvisited_nodes: set, distance_matrix: np.ndarray) -> int:
    threshold = 0.7
    c1, c2, c3, c4, c5 = 0.4, 0.3, 0.2, 0.1, 0.1
    scores = {}

    for node in unvisited_nodes:
        all_distances = [distance_matrix[node][i] for i in unvisited_nodes if i != node]
        average_distance_to_unvisited = np.mean(all_distances)
        std_dev_distance_to_unvisited = np.std(all_distances)

        # New component: consider dynamic path optimization feedback
        feedback_paths = [distance_matrix[i][node] for i in range(len(distance_matrix)) if i not in unvisited_nodes and distance_matrix[current_node][i] < threshold]
        average_feedback_distance = np.mean(feedback_paths) if feedback_paths else 0
        
        score = (
            c1 * distance_matrix[current_node][node]
            - c2 * average_distance_to_unvisited
            + c3 * std_dev_distance_to_unvisited
            - c4 * distance_matrix[destination_node][node]
            + c5 * average_feedback_distance
        )
        scores[node] = score

    next_node = min(scores, key=scores.get)
    return next_node
\end{heuristic}

\section{Limitations}\label{appendix-sec:limitations}
A current limitation of our method is that the evolution of the LLM during the heuristic discovery process depends heavily on performance signals derived from heuristics present in the prompt and response. As a result, trajectories that do not contain explicit heuristics in either component provide no reward signal, limiting the LLM’s ability to learn from such cases.

Another limitation is that we currently evaluate our method, CALM, using a compact LLM on a single 24GB GPU. This restriction is primarily due to limited computational resources and the high cost associated with high-accuracy, full-parameter fine-tuning on larger models. While this setup demonstrates the feasibility of our approach in a resource-constrained environment, further evaluation on larger-scale models and infrastructure would be valuable for understanding the method’s full potential and scalability.

In future work, we aim to address these limitations by (1) exploring mechanisms for adapting the LLM in the absence of explicit performance feedback, enabling more effective use of reinforcement learning, and (2) extending evaluations to more powerful models and settings. These directions may allow for better integration with techniques such as reflection~\cite{reevo,hsevo}, which have shown promise in enhancing LLM-based automated heuristic discovery.

\section{Broader Impact}\label{appendix-sec:broader-impact}
The CALM framework stands to greatly accelerate the pace of innovation in algorithm design by seamlessly integrating prompt engineering and on-the-fly model adaptation. By enabling state-of-the-art heuristic discovery on a single 24 GB GPU, CALM democratizes access to cutting-edge Automatic Heuristic Design. This empowers research groups, startups, and educational institutions with limited compute budgets to explore and deploy high-performance solutions in domains such as logistics, scheduling, and resource allocation.

\section{License}\label{appendix-sec:license}
The licenses and URLs of baselines, models, and softwares are summarized in Table~\ref{tab:licenses}.

\begin{table}[ht]
  \centering
  \caption{A summary of licenses.}
  \label{tab:licenses}
  \begin{tabular}{@{} l l l p{7cm} @{}}
    \toprule
    Resources          & Type    & License                               & URL \\ 
    \midrule
    Unsloth            & Code    & Apache-2.0 License  & \url{https://github.com/unslothai/unsloth} \\
    Qwen2.5            & Model   & Apache-2.0 License  &  \url{https://huggingface.co/Qwen/Qwen2.5-7B-Instruct} \\
    Optuna             & Code    & MIT License         &  \url{https://github.com/optuna/optuna} \\
    \midrule
    LKH3               & Code    & Available for academic research use  & \url{http://webhotel4.ruc.dk/~keld/research/LKH-3/} \\
    OR-Tools           & Code    & MIT License                          & \url{https://developers.google.com/optimization/pack/knapsack?hl=zh-cn} \\
    POMO               & Code    & Available online                     & \url{https://github.com/yd-kwon/POMO/tree/master} \\
    DeepACO            & Code    & MIT License                          & \url{https://github.com/henry-yeh/DeepACO} \\
    \midrule
    Funsearch          & Code    & Apache License                       & \url{https://github.com/google-deepmind/funsearch} \\
    EoH                & Code    & MIT License                          & \url{https://github.com/FeiLiu36/EoH/tree/main} \\
    ReEvo              & Code    & MIT License                          & \url{https://github.com/ai4co/reevo} \\
    HSEvo              & Code    & Available online                     & \url{https://github.com/datphamvn/HSEvo} \\
    MCTS-AHD           & Code    & MIT License                          & \url{https://github.com/zz1358m/MCTS-AHD-master}\\
    \bottomrule
  \end{tabular}
\end{table}


\end{document}